\documentclass[11pt]{article}

\usepackage[final]{acl}

\usepackage{times}
\usepackage{latexsym}
\usepackage{multirow}
\usepackage[T1]{fontenc}

\usepackage[utf8]{inputenc}

\usepackage{graphicx} 
\usepackage{algorithm}        
\usepackage{algpseudocode}    
\usepackage{booktabs}
\usepackage{amsmath}
\usepackage{microtype}

\usepackage{inconsolata}

\usepackage{graphicx}

\algrenewcommand\algorithmicrequire{\textbf{Require:}}
\algrenewcommand\algorithmicensure{\textbf{Output:}}

\title{IESR: Efficient MCTS-Based Modular Reasoning for Text-to-SQL with Large Language Models}

\author{
 \textbf{Tao Liu\textsuperscript{13*}},
 \textbf{Jiafan Lu\textsuperscript{2*}},
 \textbf{Bohan Yu\textsuperscript{3}},
 \textbf{Pengcheng Wu\textsuperscript{4}},
 \textbf{LiuHaixin\textsuperscript{3}},
 \textbf{Guoyu Xu\textsuperscript{3}},
 \textbf{lixiangheng\textsuperscript{3}},
 \\
 \textbf{Lixiao Li\textsuperscript{3}},
 \textbf{Jiaming Hou\textsuperscript{3}},
 \textbf{Zhaoshijun\textsuperscript{3}},
 \textbf{Xinglin Lyu\textsuperscript{3}},
 \textbf{Kunli Zhang\textsuperscript{3}},
 \textbf{Yuxiang Jia\textsuperscript{3}},
 \\
 \textbf{Hongyin Zan\textsuperscript{3$\dagger$}}
\\
\\
 \textsuperscript{1}Zhejiang University,
 \textsuperscript{2}Tianjin University,
 \textsuperscript{3}Zhengzhou University,
 \textsuperscript{4}Zhongshan University
\\
 {
    \href{mailto:funkytao2000@gmail.com}{funkytao2000@gmail.com},
    \href{mailto:2024244073@tju.edu.cn}{2024244073@tju.edu.cn},
    \href{mailto:iehyzan@zzu.edu.cn}{iehyzan@zzu.edu.cn}
 }
}

\begin{document}
\maketitle

\begingroup
\renewcommand\thefootnote{}
\footnotetext{*Equal contribution, $\dagger$Corresponding authors.}
\footnotetext{\textbf{Code: }\url{https://github.com/Ffunkytao/IESR-SLM}.}
\endgroup

\begin{abstract}
 Text-to-SQL is a key natural language processing task that maps natural language questions to SQL queries, enabling intuitive interaction with web-based databases. Although current methods perform well on benchmarks like BIRD and Spider, they struggle with complex reasoning, domain knowledge, and hypothetical queries, and remain costly in enterprise deployment. To address these issues, we propose a framework named \textbf{IESR}(\textbf{I}nformation \textbf{E}nhanced \textbf{S}tructured \textbf{R}easoning) for moderate-scale Large Language Models: (i) leverages Large Language Models (LLMs) for key information understanding and schema linking, and decoupling mathematical computation and SQL generation, (ii) integrates a multi-path reasoning mechanism based on Monte Carlo Tree Search (MCTS) with majority voting, and (iii) introduces a trajectory consistency verification module with a discriminator model to improve accuracy and consistency. Experimental results demonstrate that \textbf{IESR} achieves SOTA under the XiYanSQL setting on the complex reasoning benchmark LogicCat (24.28 EX) and the Archer dataset (37.28 EX), using only compact moderate-scale models without fine-tuning. Furthermore, our analysis reveals that current coder models exhibit notable biases and deficiencies in physical knowledge, mathematical computation, and common-sense reasoning, highlighting important directions for future research. We release the code in an github repository for reproducibility.
\end{abstract}

\section{Introduction}

\begin{figure}[t]
    \centering
    \includegraphics[width=\linewidth]{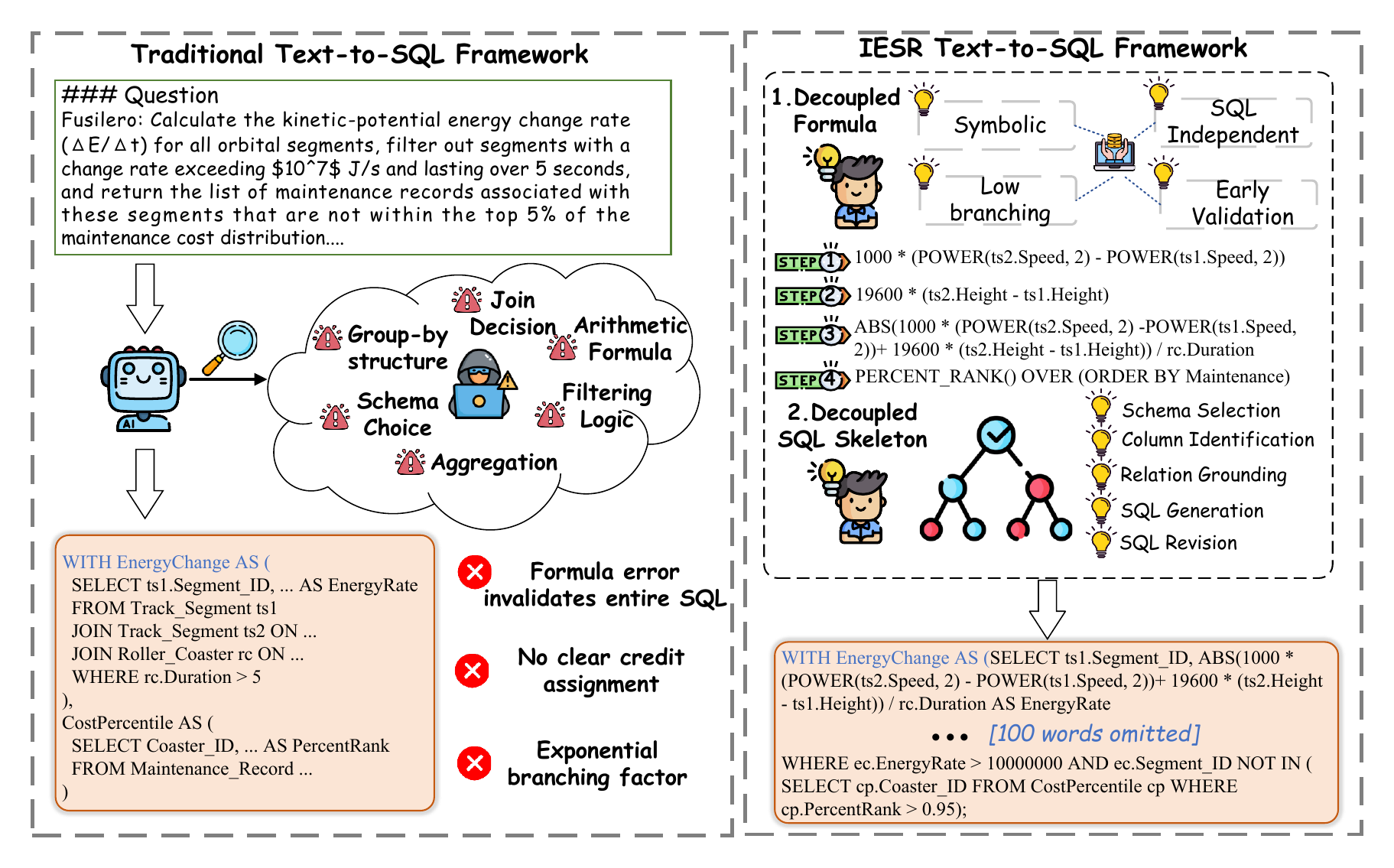}
    \caption{Motivation for Decoupling Mathematical Computation and SQL Generation in Text-to-SQL.}  
    \label{fig:figure_motiva}
\end{figure}

Text-to-SQL aims to translate natural language queries into executable SQL, enabling intuitive interaction with databases and reducing manual effort~\cite{Zelle_Mooney_1996,lei2025spider}. Recent advances have reported strong performance on widely used benchmarks such as Spider~\cite{yu2019spiderlargescalehumanlabeleddataset} and BIRD~\cite{wretblad2024understandingeffectsnoisetexttosql}, achieving execution accuracies above 90\% and 70\% respectively~\cite{pourreza2025chasesql,talaei2024chesscontextualharnessingefficient,li2025omnisqlsynthesizinghighqualitytexttosql}. However, these benchmarks largely reflect simplified settings, where many queries can be solved through shallow pattern matching and limited reasoning. In contrast, recent benchmarks including LogicCat~\cite{liu2025logiccatchainofthoughttexttosqlbenchmark} and Archer~\cite{zheng-etal-2024-archer} reveal fundamental limitations of existing Text-to-SQL systems by requiring cross-domain reasoning that integrates mathematical computation, physical units, commonsense constraints, and hypothetical conditions. Under such settings, current methods either degrade sharply or depend on substantially larger models and inference budgets, exposing a flawed assumption that schema grounding, logical reasoning, and numerical computation can be reliably resolved within a single or locally consistent generation pass. Some work~\cite{xu2025mtirsqlmultiturntoolintegratedreasoning,yao2026arctictext2sqlr1simplerewardsstrong,weng-etal-2025-graph} explores reinforcement learning and test-time optimization for Text-to-SQL using execution feedback, heuristic rewards, or consistency signals. While improving correctness, these methods often rely on heavy rollouts or learned critics and treat numerical computation as implicit, making rewards coarse in math-intensive queries and unstable under tight budgets.

Multi-agent and iterative reasoning have been explored to address this issue~\cite{talaei2024chesscontextualharnessingefficient,pourreza2025chasesql,deng2025reforcetexttosqlagentselfrefinement,feng-etal-2026-trust}, but general-purpose agent and planning paradigms~\cite{Wei_Wang_Schuurmans_Bosma_Chi_Le_Zhou,wang2023planandsolvepromptingimprovingzeroshot,shinn2023reflexion,yao2023reactsynergizingreasoningacting,song2025joltsqljointlosstuning} remain ill-suited to the structured and high-stakes nature of SQL execution. In particular, heterogeneous decision factors—such as schema selection, join construction, aggregation, filtering logic, grouping structure, and arithmetic formulas—are often tightly entangled during generation, so that minor deviations in numerical or semantic reasoning can invalidate an otherwise correct SQL structure. This tight coupling leads to unclear credit assignment, error propagation, and instability under multi-step reasoning~\cite{xia2024agentlessdemystifyingllmbasedsoftware,feng2026kbsdknowledgeboundaryaware}.

To address these limitations, we propose \textbf{IESR} (\textbf{I}nformation \textbf{E}nhanced 
\textbf{S}tructured \textbf{R}easoning) for moderate-scale Large Language Models, a modular reasoning framework for complex Text-to-SQL generation and hereafter referred to as IESR. Unlike previous search-based or agent-based works~\cite{li2025alphasql,10.1609/aaai.v39i1.31979,wang2025autolinkautonomousschemaexploration} that still treat mathematical reasoning as part of SQL generation, our formulation explicitly separates symbolic computation as an independent reasoning dimension. As illustrated in Figure~\ref{fig:figure_motiva}, IESR is motivated by the observation that mathematical computation is structurally orthogonal to SQL construction once relevant numerical attributes are identified. Accordingly, IESR decomposes the task into three tightly coupled stages that explicitly compensate for the failure modes of moderate-scale large language models: (i) an \emph{information understanding} stage that extracts semantic hypotheses and performs schema-guided compression, (ii) an \emph{MCTS-based reasoning} stage that explores multiple SQL generation trajectories with decoupled reasoning dimensions, and (iii) a \emph{trajectory selection} stage that verifies and aggregates candidate SQL paths.

Without instruction fine-tuning, IESR operates entirely with moderate-scale open-source large language models and achieves strong performance on complex mathematical, physical, and hypothetical queries. Experiments demonstrate that IESR improves reasoning-oriented Text-to-SQL performance under fixed inference-time search budgets, with the accuracy-cost trade-off controlled by the rollout number. Ablation studies further confirm the importance of information understanding, structured search, and consistency-based verification in improving robustness and execution accuracy.


The main contributions of this work are summarized as follows:
\begin{itemize}\setlength{\itemsep}{1pt}\setlength{\parskip}{0pt}
\item We propose \textbf{IESR}, a structured MCTS-based reasoning framework that integrates information understanding, schema linking, and trajectory-level verification for complex Text-to-SQL tasks.
\item We demonstrates strong performance under moderate-scale LLMs, no-finetuning settings on reasoning-oriented benchmarks.
\item We show that \textbf{IESR} is compatible with moderate-scale open-source large language models and requires no instruction fine-tuning, enabling robust Text-to-SQL reasoning under moderate-scale backbone settings.
\end{itemize}

\begin{figure*}[h]
  \centering
  \includegraphics[width=1.0\textwidth]{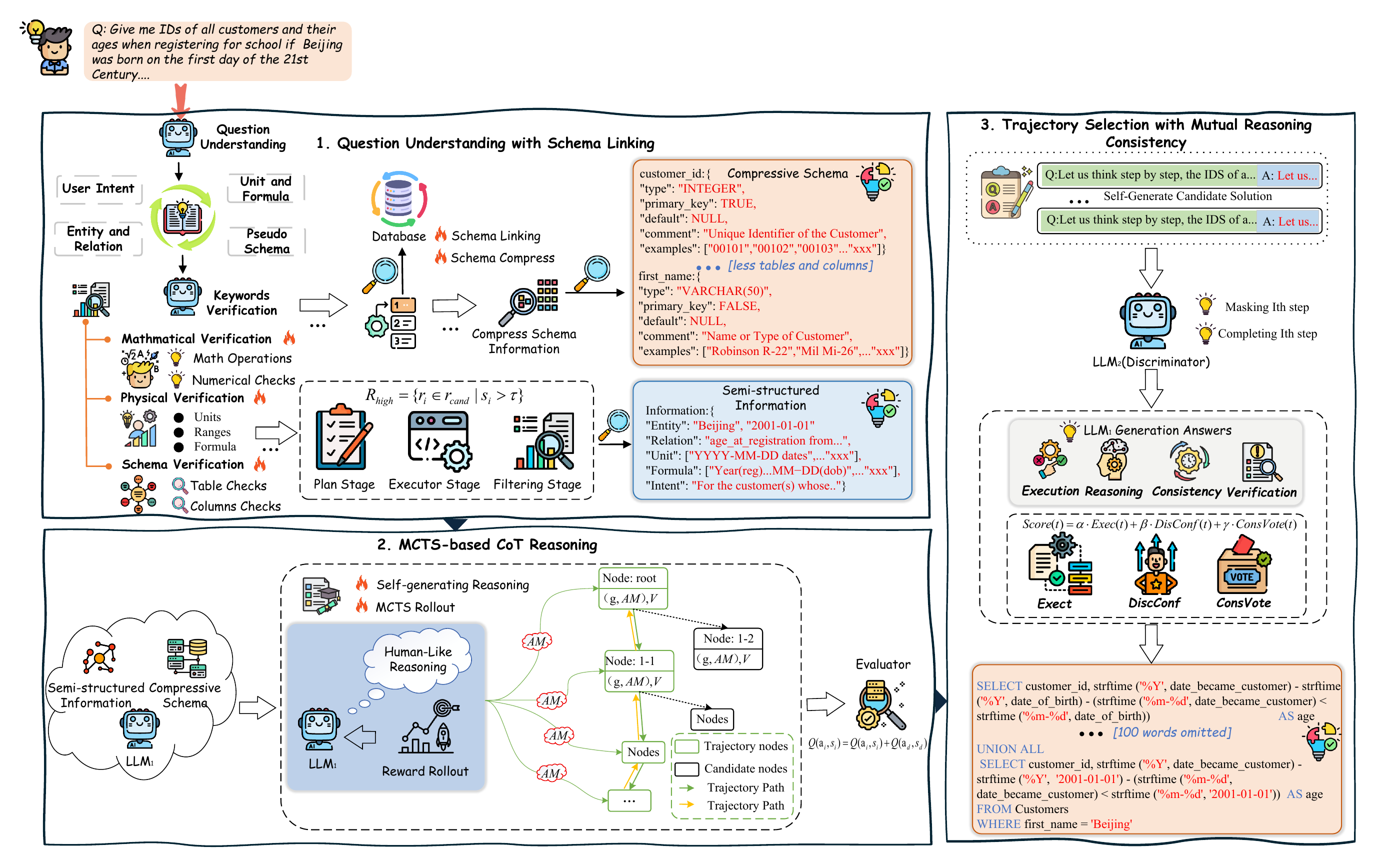}
  \caption{The comprehensive workflow of IESR including three stages: Question Understanding with Schema Linking, Monte Carlo Tree Search(MCTS)-based Reasoning and Trajectory Selection with Mutual Reasoning Consistency.}
  \label{fig:mainmodel}
\end{figure*}

\section{Related Work}

\subsection{Text-to-SQL with Decomposition and Search-based Reasoning}

Recent advances in large language models (LLMs) have substantially improved Text-to-SQL performance on benchmarks such as Spider and BIRD. Many methods decompose the task into candidate generation, refinement, and selection stages to better handle schema alignment and complex query structures~\cite{Li_2024,cao2024rslsqlrobustschemalinking,pourreza2025chasesql}. Beyond static decomposition, recent work introduces explicit search to improve exploration and robustness. Alpha-SQL~\cite{li2025alphasql} formulates Text-to-SQL as a value-guided tree search problem, while SQL-o1~\cite{lyu2025sqlo1selfrewardheuristicdynamic} and MCTS-SQL~\cite{yuan2025mctssqllightweightllmsmaster} incorporate self-reward heuristics and Monte Carlo Tree Search to guide structured exploration. Related approaches further exploit multi-trajectory reasoning or mutual verification to enhance robustness~\cite{qi2025mutual}.
IESR is related to Alpha-SQL~\cite{li2025alphasql} in using MCTS for Text-to-SQL, but differs in the search unit. Alpha-SQL mainly explores SQL trajectories with execution-derived rewards, whereas IESR searches over heterogeneous reasoning actions that separately model schema grounding, entity linking, numerical reasoning, SQL generation, and revision. This explicit action decomposition is designed for queries where numerical dependencies and domain constraints cannot be reliably handled by locally scored SQL trajectories.

\subsection{Structured Reasoning and Optimization for Text-to-SQL}

Recent work further improves Text-to-SQL via structured reasoning and optimization at training or inference time, targeting semantic faithfulness under limited computational budgets. Multi-turn and tool-integrated frameworks, such as MTIR-SQL~\cite{xu2025mtirsqlmultiturntoolintegratedreasoning}, optimize generation trajectories through reinforcement signals, while Arctic-Text2SQL-R1~\cite{yao2025arctic} shows that simple reward designs can already induce strong reasoning behaviors~\cite{yao2026arctictext2sqlr1simplerewardsstrong}. To reduce execution overhead, Graph-Reward-SQL introduces execution-free reward modeling via graph matching and stepwise supervision~\cite{weng-etal-2025-graph}. In parallel, retrieval-aware optimization mitigates schema context cost: in-context reinforcement learning with retrieval-augmented generation improves database and table selection under constrained context windows~\cite{toteja-etal-2025-context}. Test-time scaling strategies further allocate inference budgets to improve robustness, as explored in Agentar-Scale-SQL~\cite{wang2025agentarscalesqladvancingtexttosqlorchestrated} and SERL-SQL~\cite{liu2026serlsqlselectivehindsightdistillation}. IESR also differs from CHASE-SQL~\cite{pourreza2025chasesql} and DeepEye-SQL~\cite{li2025deepeyesqlsoftwareengineeringinspiredtexttosqlframework}. CHASE-SQL focuses on multi-path candidate generation and preference-based final selection, while DeepEye-SQL emphasizes schema analysis and SQL refinement. In contrast, IESR verifies intermediate reasoning trajectories through masking-and-completion consistency and explicitly separates symbolic computation from SQL construction, making it more targeted toward math-intensive and domain-constrained Text-to-SQL.

\section{Methodology}
Motivated by the need to extract structured semantics and user intent from natural language questions and propagate them into schema linking and multi-step reasoning. Figure~\ref{fig:mainmodel} presents IESR, a modular Text-to-SQL framework with three components: question understanding, MCTS-based reasoning, and trajectory selection. IESR first extracts structured semantics and aligns them with relevant schema, then explores multiple reasoning trajectories via MCTS, and finally select the most reliable SQL.

\subsection{Question Information Understanding}

\paragraph{Intent and Information Understanding.}
To support schema selection and downstream reasoning, we first map each multi-domain query into an intermediate \emph{latent semantic state}, which encodes query-implied entities, relations, numerical cues, units, and field patterns before schema grounding. Detailed algorithm listed at Appendix~\ref{sec:algorithm}.

Given a query $q$ and schema $S=(T,C)$, we use a moderate-scale LLM to produce a latent semantic state
$\mathcal{S}_q=(i,r,E,\mathcal{R},N,U,P)$, where $E$ denotes extracted entities, $\mathcal{R}$ denotes candidate relations and schema references, $N$ denotes numeric expressions, $U$ denotes associated units, and $P$ denotes candidate field patterns. We treat $\mathcal{S}_q$ as a high-recall set of noisy semantic hypotheses, which is later refined through consistency verification to guide schema compression and reasoning.



\paragraph{Constraint-aware Relation Filtering.}
To mitigate hallucination and semantic drift, we introduce a set of domain-agnostic \emph{consistency constraints} that evaluate the internal coherence of semantic hypotheses. From $\mathcal{S}_q$, we obtain an initial relation set $\mathcal{R}_{\text{init}}$. Each constraint encodes compatibility among entities, units, and transformation patterns as
$C_j = (\text{Enti}_j, \text{Uni}_j, \text{Equt}_j)$.


Candidate relations are filtered via semantic compatibility:
\begin{equation}
\small
\mathcal{R}_{\text{cand}}
=
\{\, r_i \in \mathcal{R}_{\text{init}} \mid
\textsc{Match}(r_i, \mathcal{C}) \,\}.
\end{equation}
Here, $\textsc{Match}(r_i, \mathcal{C})$ holds if there exists a constraint
$C_j \in \mathcal{C}$ such that
$\text{sim}(r_i, \text{Uni}_j, \text{Equt}_j) > \delta_{\text{match}}$ and $\text{sim}(\cdot)$ is implemented by a matcher $\mathcal{M}_{\text{match}}$.

\paragraph{Soft Consistency Scoring.}
The filtered relations $\mathcal{R}_{\text{cand}}$ are evaluated by a \emph{Plan \& Executor} module, which assigns a soft consistency score:
\begin{equation}
\small
P_i = \text{Plan}(r_i, \mathcal{C}), \quad
s_i = \text{Executor}(r_i, P_i),
\end{equation}
where $s_i \in [0,1]$ reflects the degree of consistency between a semantic hypothesis and the constraint set. We intentionally adopt this non-learned, constraint-based scorer to ensure robustness in low-resource settings and avoid additional supervision or training overhead.

Relations with scores above a threshold $\tau$ are retained:
\begin{equation}
\small
\mathcal{R}_{\text{high}} = \{\, r_i \in \mathcal{R}_{\text{cand}} \mid s_i > \tau \,\}.
\end{equation}

The resulting validated semantic state conditions subsequent schema linking and compression, enabling relevance estimation beyond the raw query.

\paragraph{Schema Linking and Compression.}

Conditioned on the validated semantic state from the previous stage, we perform schema linking and compression to surface task-relevant structures and constrain the reasoning search space. Following the M-Schema technique~\cite{XiYanSQL}, we encode schema elements in a semi-structured format with explicit annotations, and apply filtering to retain only high-salience fields.
We employ two complementary strategies:
\begin{itemize}
    \item \textbf{Locality-Sensitive Hashing:} filters tables and columns by lexical similarity, efficiently narrowing candidate schema elements.
    \item \textbf{Semantic Similarity Matching:} selects schema elements whose semantic representations are most relevant to the conditioned query.
\end{itemize}
Given the large size of full schema where only a small subset is relevant per query, the refined schema with key fields and annotations is provided to the reasoning agent for subsequent processing.

\subsection{MCTS-based CoT Reasoning}



Following prior work~\cite{qi2025mutual,li2025alphasql}, we cast Text-to-SQL generation as structured search over multi-step reasoning trajectories. Complex SQL requires interdependent decisions over schema grounding, joins, aggregation, filters, and numerical computation, which moderate-scale LLMs often cannot resolve reliably in a single pass. MCTS addresses this by representing each node as a reasoning state with a partial SQL hypothesis and semantic context, each edge as a reasoning action, and each root-to-terminal path as a candidate trajectory. Its exploration-exploitation tradeoff allows promising partial solutions to be expanded while preserving alternative paths, making it more suitable than beam search or self-consistency for heterogeneous reasoning actions. After search, terminal trajectories are collected as $\mathcal{T}=\{t_1,t_2,\ldots,t_n\}$, where $n\geq1$, and passed to the trajectory selection module. The detailed algorithm is provided in Appendix~\ref{sec:algorithm}.

\paragraph{Human-inspired Reasoning Actions.}

\begin{figure}[t]
  \centering
  \includegraphics[width=0.5\textwidth]{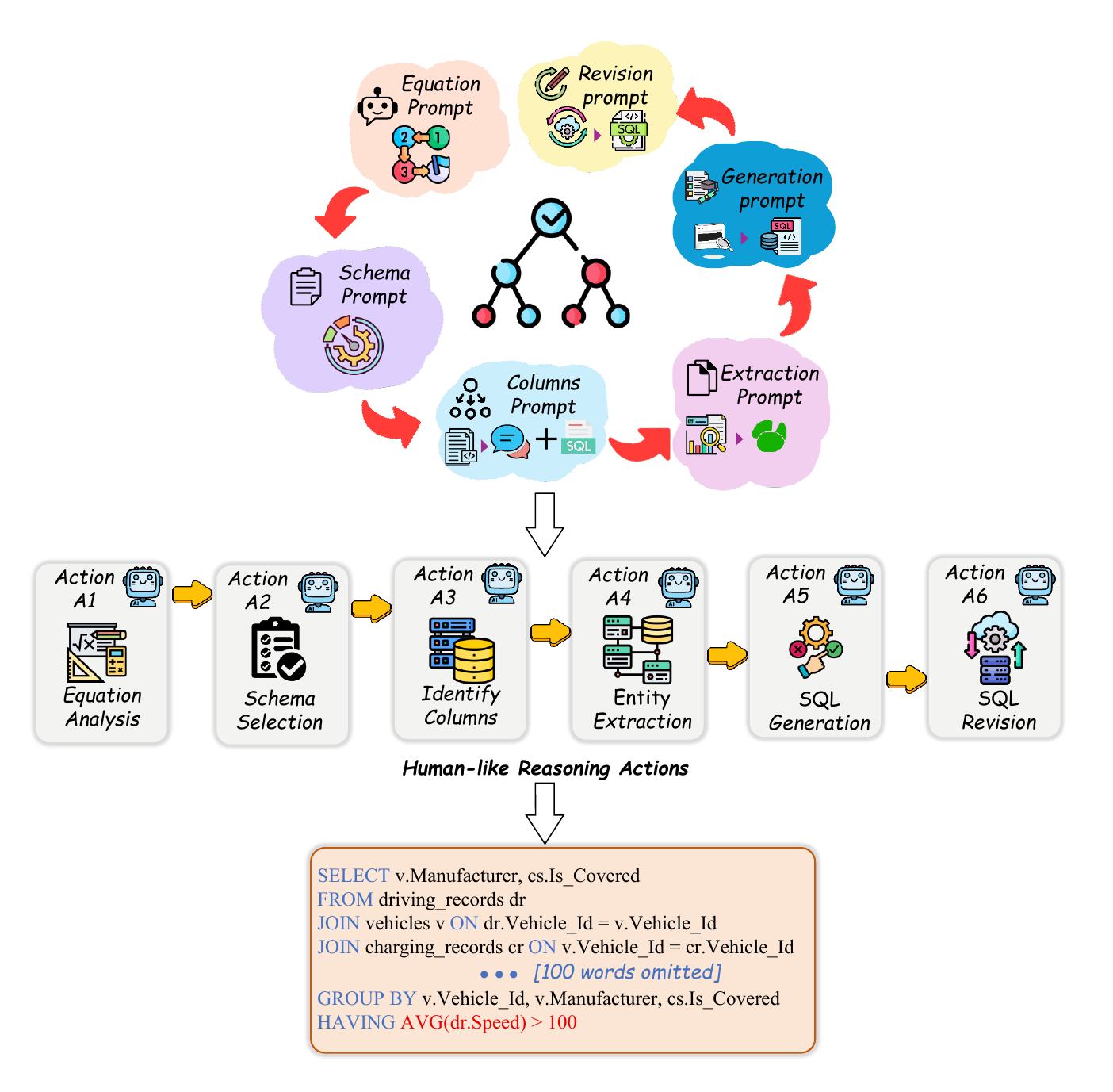}
  \caption{A visual illustration of heterogeneous MCTS actions (A1–A6) for SQL generation and reasoning.}
  \label{fig:ActionA15}
\end{figure}

Most MCTS-based reasoning methods use a single action type, usually next-step generation~\cite{chen2024alphamathzeroprocesssupervision}. For complex SQL, this coarse transition is inefficient because schema grounding, numerical reasoning, and structural refinement require different operations. Each reasoning action is defined as a state transition over the partial SQL structure and semantic context. Guided by the problem-solving patterns in Figure~\ref{fig:ActionA15}, we design a heterogeneous action space for dimension-specific exploration. Each action updates the reasoning state along a specific semantic dimension, enabling more targeted exploration than generic next-step expansion. The formal node representation is provided in Appendix~\ref{sec:nodes}.

\begin{itemize}\setlength{\itemsep}{1pt}\setlength{\parskip}{0pt}
    \item \textbf{Equation Analysis}: explicit modeling of formulas and numerical relations~\cite{mao2025steinersqlgraphguidedmathematicalreasoning}.
    \item \textbf{Schema Selection}: identifying task-relevant tables and columns from large schemas~\cite{li2025alphasql,talaei2024chesscontextualharnessingefficient}.
    \item \textbf{Identify Columns}: resolving ambiguous schema fields using semi-structured semantic cues~\cite{li2025deepeyesqlsoftwareengineeringinspiredtexttosqlframework}.
    \item \textbf{Entity Extraction}: grounding entities and relations for accurate filtering and joins.
    \item \textbf{SQL Generation}: constructing executable SQL queries, particularly nested structures~\cite{pourreza2025chasesql}.
    \item \textbf{SQL Revision}: correcting erroneous queries via reasoning over formulas, units, and domain knowledge.
\end{itemize}

At each step, MCTS selects an action from this space and generates the next reasoning state accordingly.


\paragraph{Reward-based Node Evaluation.}
The reward function guides trajectory quality in MCTS. In our low-resource setting, we adopt an execution-based, terminal-only self-consistency reward without learned critics, following Alpha-SQL~\cite{li2025alphasql}. 
When MCTS reaches a terminal node and produces a candidate SQL $y$, we sample $N$ additional SQL queries $\{y_i\}_{i=1}^{N}$ under the same terminal context and execute them on database $D$. The reward is defined as the agreement rate of execution results:
\begin{equation}
\small
\begin{aligned}
r_t \;=\; R(y,q,D)
&= \frac{1}{N}\sum_{i=1}^{N}
\mathbf{1}\Big[
\textsc{Execute}(y,D) \\
&\qquad\qquad = \textsc{Execute}(y_i,D)
\Big].
\end{aligned}
\end{equation}

\paragraph{MCTS Search and Backpropagation.}
Given a rollout path consisting of nodes $v_0,\ldots,v_d$ and actions $a_1,\ldots,a_d$, we backpropagate the terminal reward $r_t$ to update visit counts and cumulative values:
\begin{equation}
\small
\begin{aligned}
N(v_i, a_{i+1}) &\leftarrow N(v_i, a_{i+1}) + 1, \\
Q(v_i, a_{i+1}) &\leftarrow Q(v_i, a_{i+1}) + R(y,q,D),
\end{aligned}
\end{equation}
for $i=0,\ldots,d-1$. Action selection follows the standard UCT criterion:
\begin{equation}
\small
UCT(v,a) = \frac{Q(v,a)}{N(v,a)} + c \sqrt{\frac{\ln N(v)}{N(v,a)}}.
\end{equation}
The search iterates through selection, expansion, simulation, and backpropagation. After $N_{\text{rollout}}$ rollouts, all terminal trajectories are collected as candidate SQL queries for subsequent selection and verification.

\subsection{Trajectory Selection with Mutual Reasoning Consistency}

To reliably select executable and semantically consistent SQL from noisy candidates, we introduce a collaborative selection mechanism between a primary model $LLM_{1}$ and a secondary verifier $LLM_{2}$. Detailed algorithm listed at Appendix~\ref{sec:algorithm}.

\paragraph{Discriminator Consistency Verification.}
For a candidate reasoning trajectory $t = x \oplus s_{1} \oplus s_{2} \oplus \cdots \oplus s_{d}$, we assess its logical self-consistency via a masking-and-completion procedure. Specifically, we mask the reasoning process at an intermediate step $i$ ($i < d$), retaining only the first $i-1$ intermediate steps such as selected tables, partial columns, or partial \texttt{WHERE} conditions. The incomplete sequence is then provided to the verifier $LLM_{2}$ for completion.


If the completion produced by $LLM_{2}$ is semantically consistent with the original trajectory $t$, yielding compatible schema grounding and SQL structure, we regard the trajectory as stable and retain it. This masking-and-completion check tests whether the same semantic constraints can be recovered from partial reasoning states. $LLM_{2}$ is a frozen verifier with scale comparable to $LLM_{1}$ and requires no additional supervision.

\paragraph{Scoring Mechanism.}
To select the most reliable SQL from filtered candidates, we assign a composite score to each trajectory:
\begin{equation}
\small
\begin{aligned}
\text{Score}(t)
&= \alpha \cdot \text{Exec}(t) \\
&\quad + \beta \cdot \text{DiscConf}(t)
 + \gamma \cdot \text{ConsVote}(t).
\end{aligned}
\end{equation}

Here, $\text{Exec}(t)$ evaluates execution correctness or equivalence when available, while $\text{DiscConf}(t)$ and $\text{ConsVote}(t)$ provide consistency signals derived from discriminator verification and peer agreement, respectively. The coefficients $\alpha, \beta, \gamma$ are fixed to balance executability and consistency, with consistency serving as auxiliary evidence. Detailed representation is illustrated at Appendix~\ref{sec:weight_analysis}. The final SQL answer is selected by maximizing $\text{Score}(t)$ over all candidate trajectories.

\section{Experiments}
We conduct comprehensive experiments to evaluate effectiveness, robustness, and inference-budget trade-offs on complex Text-to-SQL benchmarks, with a particular focus on mathematical, physical, and hypothetical reasoning under model settings.
\subsection{Experimental Setup}

\noindent\textbf{Datasets.}
We evaluate IESR on two challenging reasoning-oriented benchmarks, LogicCat~\cite{liu2025logiccatchainofthoughttexttosqlbenchmark} and Archer~\cite{zheng-etal-2024-archer}, which require multi-domain reasoning including mathematics, physics, commonsense, and hypothetical analysis. 
Detailed datasets listed at Appendix~\ref{sec:dataset}.

\begin{table*}[htb]
\centering
\renewcommand{\arraystretch}{0.92}
\setlength{\tabcolsep}{0.8mm}
\footnotesize
\begin{tabular}{l|cccc|cccc}
\toprule
\textbf{Method} & \textbf{Qwen2.5-Coder} & \textbf{XiYanSQL} & \multicolumn{1}{c|}{\textbf{OmniSQL}} & \textbf{Seed-Coder} & \textbf{Qwen2.5-Coder} & \textbf{XiYanSQL} & \multicolumn{1}{c|}{\textbf{OmniSQL}} & \textbf{Seed-Coder} \\
\midrule
\textbf{Size}
& \multicolumn{3}{c|}{7B} & 8B
& \multicolumn{3}{c|}{7B} & 8B \\
\midrule
\multicolumn{1}{c|}{}
& \multicolumn{4}{c|}{\textbf{LogicCat EX(\%)}}
& \multicolumn{4}{c}{\textbf{Archer-Dev EX(\%)}} \\
\midrule
Cot-SQL      & 5.07  & 7.89  & 6.93  & 6.58  & 17.16 & 18.46 & 18.14 & 19.89 \\
DIN-SQL      & 10.31 & 11.56 & 9.56  & 8.98  & 31.45 & 34.41 & 26.04 & 29.13 \\
DAIL-SQL     & 10.73 & 10.98 & 9.43  & 8.18  & 29.67 & 32.28 & 27.56 & 24.12 \\
DTS-SQL      & 14.12 & 14.88 & 12.67 & 10.31 & 31.88 & 33.17 & 32.14 & 30.33 \\
CHESS-SQL    & 16.16 & 19.21 & 10.89 & 14.52 & 32.45 & 35.54 & 27.27 & \textbf{31.56} \\
Chase-SQL    & 21.92 & 22.13 & 19.76 & 21.56 & \textbf{36.12} & 35.78 & 27.12 & 28.98 \\
LinkAlign    & 19.56 & 22.56 & 14.34 & 18.89 & 29.21 & 31.28 & 30.21 & 30.10 \\
Alpha-SQL    & 18.32 & 19.78 & 16.56 & 19.34 & 28.32 & 30.12 & 29.88 & 31.22 \\
Deepeye-SQL  & \textbf{22.81} & 21.22 & 22.30 & 21.75 & 19.23 & 21.15 & 17.12 & 13.46 \\
SQL-O1       & 16.42 & 17.91 & 13.24 & 17.33 & 26.56 & 28.22 & 27.41 & 28.78 \\
Standard CoT & 10.07 & 16.83 & 11.56 & 14.53 & - & - & - & - \\
\textbf{IESR}
& 21.66 & \textbf{24.28} & \textbf{22.41} & \textbf{22.11}
& 35.21 & \textbf{37.28} & \textbf{33.11} & 30.66 \\
\bottomrule
\end{tabular}
\caption{Main results on reasoning-intensive Text-to-SQL benchmarks. EX denotes execution accuracy. Bold values indicate the best result in each backbone-dataset column. SQL-O1 requires task-specific tuning.}
\label{tab:reasoning_results_ex}
\end{table*}

\noindent\textbf{Baseline.}
We compare IESR with representative SOTA Text-to-SQL systems, including DIN-SQL~\cite{pourreza2023dinsql}, DAIL-SQL~\cite{gao2023texttosqlempoweredlargelanguage}, DTS-SQL~\cite{pourreza-rafiei-2024-dts}, CHESS~\cite{talaei2024chesscontextualharnessingefficient}, LinkAlign~\cite{wang2025linkalignscalableschemalinking}, Alpha-SQL~\cite{li2025alphasql}, Chase-SQL~\cite{pourreza2025chasesql}, Deepeye-SQL~\cite{li2025deepeyesqlsoftwareengineeringinspiredtexttosqlframework} and SQL-O1~\cite{lyu2025sqlo1selfrewardheuristicdynamic}. These methods cover a broad range of paradigms such as decomposed prompting, schema linking, hierarchical reasoning, and MCTS-based search.  
Detailed public methods listed at Appendix~\ref{sec:dataset}.

\noindent\textbf{Backbone Models.}
For the information understanding module, we evaluate several general-purpose LLMs and select Qwen3-8B~\cite{yang2025qwen3technicalreport}  for its favorable balance between reasoning capability and efficiency.  
For MCTS-based reasoning, we adopt four code-oriented models: Qwen2.5-Coder-7B~\cite{hui2024qwen25codertechnicalreport}, XiYanSQL-QwenCoder-7B-2504~\cite{gao2025previewxiyansqlmultigeneratorensemble}, Seed-Coder-8B~\cite{seed2025seedcoderletcodemodel}, and OmniSQL-7B~\cite{li2025omnisqlsynthesizinghighqualitytexttosql}.   
For trajectory consistency verification, we employ Qwen3-4B as a discriminator, which offers strong consistency assessment with minimal overhead.

\noindent\textbf{Evaluation Metric.}
We adopt the widely used metric Execution Accuracy (EX)~\cite{pourreza2025chasesql}. This evaluates where a query is correct only if its execution result matches that of the ground-truth query.

\noindent\textbf{Experiment Details.}
Experiments were conducted on a cloud server with four RTX 4090 GPUs. Local LLM inference was accelerated with vLLM. Baseline hyperparameters follow their original implementations, while the detailed MCTS and decoding configurations are reported in Appendix~\ref{sec:impl_config}, and all prompts are provided in Appendix~\ref{sec:prompt}.

\begin{table*}[t]
\centering
\small
\renewcommand{\arraystretch}{0.92} 
\begin{tabular}{l cc cc cc cc}
\toprule
& \multicolumn{2}{c}{\textbf{Qwen2.5-Coder-7B}} & \multicolumn{2}{c}{\textbf{XiYanSQL-QwenCoder-7B}} & \multicolumn{2}{c}{\textbf{Seed-coder-8B}} & \multicolumn{2}{c}{\textbf{OmniSQL-7B}} \\
\cmidrule(lr){2-3} \cmidrule(lr){4-5} \cmidrule(lr){6-7} \cmidrule(lr){8-9}
\textbf{Configuration} & \textbf{EX} & \textbf{Drop} & \textbf{EX} & \textbf{Drop} & \textbf{EX} & \textbf{Drop} & \textbf{EX} & \textbf{Drop} \\
\midrule
\textbf{Full Model (IESR)} & \textbf{21.66} & - & \textbf{24.28} & - & \textbf{22.11} & - & \textbf{22.41} & - \\
\midrule
\textit{Core Stage Isolation} \\

- Stage 1.1+2+3 Only (w/o) & 19.96 & 1.70 & 21.68 & 2.60 & 20.41 & 1.70 & 21.42 & 0.99 \\
- Stage 1.2+2+3 Only (w/o) & 19.28 & 2.38 & 20.21 & 4.07 & 19.22 & 2.89 & 20.42 & 1.99 \\
- Stage 1+2+3.1 Only (w/o) & 19.24 & 2.42 & 21.83 & 2.45 & 18.29 & 3.82 & 19.14 & 3.27 \\
- Stage 1+2+3.2 Only (w/o) & 18.11 & 3.55 & 20.63 & 3.65 & 17.21 & 4.90 & 18.94 & 3.47 \\
- Stage 1 (w/o) & 11.12 & 10.54 & 14.36 & 9.92 & 9.42 & 12.69 & 12.67 & 9.74 \\

\bottomrule
\end{tabular}
\caption{Comprehensive ablation study of IESR components across four backbone models on the LogicCat dataset. We report \textbf{EX} and the absolute performance drop in percentages (\%). Understanding for Intent and Information Understanding. Linking for Schema Linking and Compression. Consistency Verification for Discriminator Consistency Verification. Reasoning and Discriminator Agent for MCTS-based CoT Reasoning and Trajectory Selection with Mutual Reasoning Consistency. }
\label{tab:comprehensive_ablation_four_backbones}
\end{table*}

\subsection{Main Results}
\noindent\textbf{Overall Performance.}
Table~\ref{tab:reasoning_results_ex} reports the main results on LogicCat and Archer, where SQL generation requires joint reasoning over schema grounding, numerical computation, physical constraints, and hypothetical conditions. IESR shows stronger performance on reasoning-intensive settings under the evaluated small-backbone configurations on these reasoning-intensive settings, indicating that explicit task decomposition, MCTS-based trajectory exploration, and consistency verification are most useful when local SQL decisions are tightly coupled with external semantic constraints. We report BIRD and Spider results in Appendix~\ref{sec:general_benchmark_results}; the weaker transfer on these conventional benchmarks suggests that IESR primarily benefits complex reasoning scenarios, while standard schema-grounding tasks may favor more specialized ranking or linking pipelines.

\begin{figure}[htb]
     \centering
     \includegraphics[width=\linewidth]{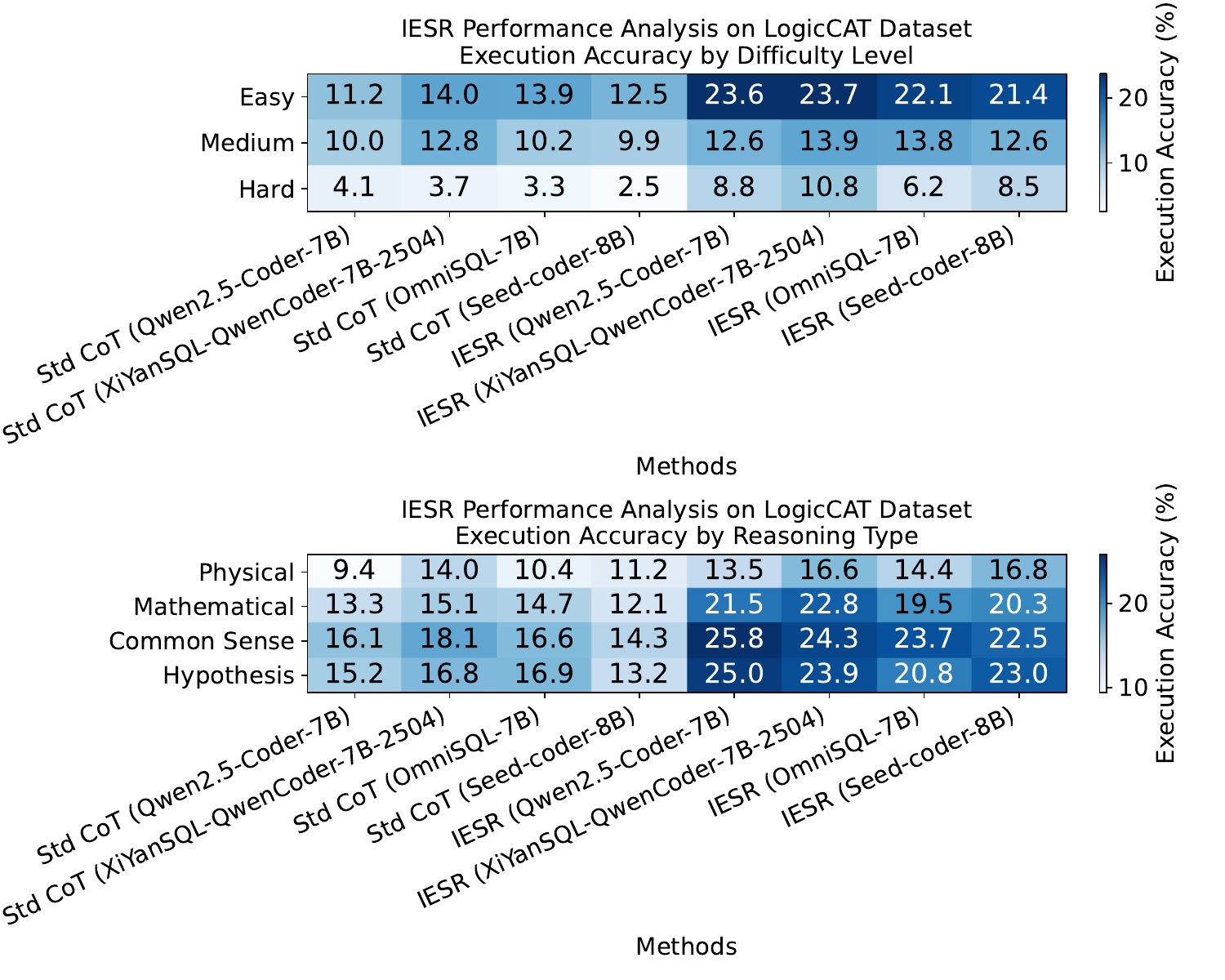}
     \caption{Performance heatmap of different methods on the LogicCat dataset across three difficulty levels (Easy, Medium, Hard) and different reasoning types.}
     \label{fig:LogicCat_heatmap}  
\end{figure}

\noindent\textbf{LogicCAT Analysis.}
As illustrated in Figure~\ref{fig:LogicCat_heatmap}, IESR achieves substantial improvements across all difficulty levels, with the largest relative gains are often observed on Hard.
Further analysis by reasoning type reveals that current code-oriented LLMs perform significantly worse on physical and mathematical reasoning compared to commonsense queries, highlighting a critical limitation. IESR effectively mitigates this gap through structured reasoning and trajectory verification. 

\subsection{Ablation Analysis}



\paragraph{Component Contribution Analysis.}
Table~\ref{tab:comprehensive_ablation_four_backbones} presents ablation results for IESR. Removing any module degrades performance, with the largest drops observed when disabling MCTS-based reasoning or schema linking, indicating their central role. Consistency verification and scoring yield smaller but consistent improvements, confirming their effectiveness as reliability enhancements. More error analysis are listed at Appendix~\ref{sec:compariAnalysis}.

\begin{table}[t]
\centering
\small
\setlength{\tabcolsep}{0.5mm}
\resizebox{\columnwidth}{!}{
\begin{tabular}{lccccc}
\toprule
\multirow{2}{*}{\textbf{Variant}}
& \multirow{2}{*}{\textbf{EX}}
& \multicolumn{4}{c}{\textbf{Diagnostics}} \\
\cmidrule(lr){3-6}
&
&
\textbf{Valid}
& \textbf{Schema}
& \textbf{Formula}
& \textbf{Unit} \\
\midrule

Coupled MCTS
& 8.00
& 37.2
& 13.0
& 86.4
& 6.5 \\

Decoupled w/o Schema
& 5.00
& 46.3
& 12.0
& 85.2
& 4.3 \\

Decoupled w/ Schema
& 11.30
& \textbf{61.0}
& 13.7
& 87.0
& 11.3 \\

Typed-Action MCTS
& 13.80
& 56.9
& 17.2
& 85.1
& 11.5 \\

Full IESR
& \textbf{21.24}
& 57.3
& \textbf{20.0}
& \textbf{91.3}
& \textbf{17.1} \\

\bottomrule
\end{tabular}
}
\caption{
Controlled comparison of reasoning formulations on LogicCat using Qwen2.5-Coder-7B-Instruct.
Decoupled denotes decomposition of formula reasoning and SQL
skeleton construction. Schema denotes table and column grounding
accuracy. Formula and Unit are evaluated on their corresponding
reasoning-sensitive subsets.
}
\label{tab:search_formulation}
\end{table}

Table~\ref{tab:search_formulation} shows that the gains of IESR come from structured interaction among reasoning components rather than decomposition alone. Decoupling without schema conditioning can produce locally plausible but weakly grounded intermediate reasoning, while schema conditioning better aligns symbolic computation with database semantics. Typed-action MCTS further separates heterogeneous decisions during search, and the full model benefits from verification to suppress locally consistent yet globally incorrect trajectories. Overall, the results support jointly structuring grounding, search, and verification for reasoning-intensive Text-to-SQL.

\begin{table}[tb]
\small
\centering
\setlength{\tabcolsep}{0.5mm}
\resizebox{\columnwidth}{!}{
\begin{tabular}{c|cccc}
\toprule
\textbf{Action} & \textbf{Qwen2.5-Coder} & \textbf{XiYanSQL} & \textbf{OmniSQL} & \textbf{Seed-Coder} \\
\midrule
All actions  
& 21.66 & 24.28 & 22.11 & 22.41 \\
w/o A1 
& 21.45 (↓ 0.21) & 24.02 (↓ 0.26) & 21.92 (↓ 0.19) & 22.23 (↓ 0.18) \\
w/o A2  
& 20.83 (↓ 0.83) & 23.88 (↓ 0.40) & 20.68 (↓ 1.43) & 22.01 (↓ 0.40) \\
w/o A3 
& 20.32 (↓ 1.34) & 23.25 (↓ 1.03) & 19.78 (↓ 0.33) & 22.12 (↓ 0.29) \\
w/o A4 
& 20.86 (↓ 0.80) & 23.87 (↓ 1.31) & 20.89 (↓ 0.80) & 21.98 (↓ 0.43) \\
w/o A6 
& 20.12 (↓ \textbf{1.54}) & 22.12 (↓ \textbf{2.16}) & 19.21 (↓ \textbf{2.90}) & 21.15 (↓ \textbf{1.26}) \\
\bottomrule
\end{tabular}
}
\caption{Ablation study on the reasoning action space (A1--A4, A6), corresponding to the reasoning actions defined in Section~3.2. We do not ablate A5 (SQL Generation) since removing it yields degenerate trajectories. The evaluation metric is \textbf{EX} (execution accuracy).}
\label{tab:action}
\end{table}

\paragraph{Action Space Analysis.}
This experiment evaluates the contribution of individual reasoning actions in the proposed action space.
Following rStar~\cite{qi2025mutual}, we conduct action-level ablation on the LogicCat dataset by removing one reasoning action at a time while keeping all other settings unchanged. The results in Table~\ref{tab:action} show that removing any remaining action consistently degrades performance across all backbone models, indicating that each reasoning action contributes to effective exploration. Notably, removing the SQL revision action results in the largest performance drop, highlighting its critical role in correcting intermediate reasoning errors and refining executable SQL queries. The varying sensitivity to individual actions across backbones suggests that action decomposition can compensate for different model-specific weaknesses, providing structured inductive bias beyond the reasoning capability of any single backbone.

\begin{table}[t]
\centering
\resizebox{\linewidth}{!}{%
\begin{tabular}{l *{8}{r}}
\toprule
  & \multicolumn{2}{c}{table}
  & \multicolumn{2}{c}{column}
  & \multicolumn{2}{c}{unit\_num}
  & \multicolumn{2}{c}{operator} \\
\cmidrule(lr){2-3}\cmidrule(lr){4-5}\cmidrule(lr){6-7}\cmidrule(lr){8-9}
 & P & R & P & R & P & R & P & R \\
\midrule
Qwen2.5-7B-Instruct      & 64.60 & 63.80 & 57.28 & 57.77 & 57.86 & 57.90 & 77.85 & 73.96 \\
Qwen3-8B                 & 75.25 & 73.54 & 63.32 & 62.30 & 60.80 & 61.06 & 77.02 & 75.36 \\
Deepseek-V3              & 76.72 & 74.81 & 67.34 & 66.67 & 62.88 & 62.18 & 78.18 & 74.03 \\
Deepseek-R1              & 78.09 & 75.35 & 62.91 & 62.02 & 60.52 & 60.32 & 76.00 & 72.20 \\
Gemini2.5-Flash          & 87.08 & 92.79 & 77.70 & 82.78 & 78.94 & 80.20 & 92.75 & \textbf{92.16 }\\
GPT-4o                   & 88.73 & \textbf{93.02} & \textbf{79.80} & 84.11 & 79.77 & 80.31 & 92.75 & 92.10 \\
Qwen2.5-7B-Instruct+Ve   & 87.46 & 89.87 & 79.41 & 82.72 & 79.88 & 80.30 & 91.53 & 89.50 \\
\textbf{Qwen3-8B+Ve}             & \textbf{90.09} & 91.24 & 78.20 & \textbf{83.24} & \textbf{81.02} & \textbf{81.11} & \textbf{92.91} & 91.31 \\
\bottomrule
\end{tabular}
}%
\caption{Precisions and Recalls of schema items, units, key numbers, and compute operators used in verification. \textbf{Ve} for Keywords Verification.}
\label{tab:verif_metrics}
\end{table}

\subsection{$N_\text{rollout}$ Analysis}
\begin{figure}[tb]
     \centering
     \includegraphics[width=\linewidth]{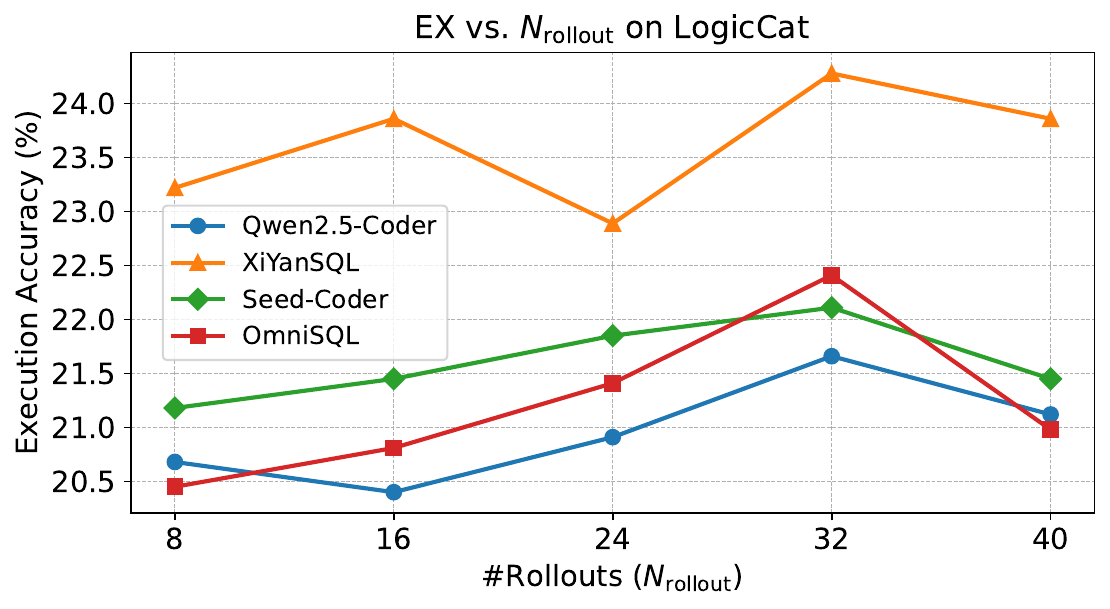}
     \caption{Ablation study of $N_\text{rollout}$ across four backbone models on the LogicCat dataset. 
     }
     \label{fig:rollouts}
\end{figure}

Shown in Figure~\ref{fig:rollouts} disabling schema compression leads to a clear performance decrease, confirming the role of schema understanding in Text-to-SQL tasks. We observe that increasing the number of $N_\text{rollout}$ generally improves the reliability and performance of the output, as greater diversity and consensus help filter out spurious results. In particular, our best results are achieved with 32 $N_\text{rollout}$, indicating that a larger ensemble improves robustness. However, as the number of $N_\text{rollout}$ increases beyond 32, performance gains tend to plateau and may even decline. This is likely because excessive rollouts introduce more noisy or low-quality reasoning paths, which can dilute the effectiveness of the majority-voting strategy. Detailed Experiments are shown at Appendix~\ref{sec:rolloutAnalysis}.

\subsection{Different Models on Information Understanding Analysis}
With the Verification module in place, shown in Table~\ref{tab:verif_metrics}, the extractor shifts from a pattern matching heuristic to a recall amplifying pipeline that systematically recovers missed schema cues and normalizes numbers and units without sacrificing precision. This manifests itself most strongly in the long-tail fields (\textit{column}, \textit{unit\_num}) where add Verification delivers large double-digit recall gains, while on \textit{table} and \textit{operator} the verified Qwen3-8B climbs into the same performance regime as frontier models such as GPT-4o and Gemini-2.5-Flash. The improvements are not cosmetic: precision rises alongside recall, indicating that verification corrects upstream omissions and inconsistencies rather than merely relaxing decision thresholds. 

\subsection{Comparison with Baseline LLMs}
We compare IESR with both general-purpose and reasoning-oriented LLMs on LogicCat under the same Text-to-SQL prompting setting. As shown in Table \ref{tab:llmcompare}, among the baseline models, o4-mini achieves the highest standalone accuracy, outperforming both general-purpose models such as GPT-4.1 and reasoning-focused models such as Gemini-2.5-Flash. However, IESR even when paired with a relatively moderate-scale large 7B parameter model, exceeds the compared zero-shot LLM prompting baselines on LogicCat under the same prompt, including o4-mini, highlighting the strength of the framework in targeting reasoning optimization. We further apply IESR to Qwen2.5-7B-Instruct and Qwen3-8B, yielding substantial gains over direct prompting and confirming that the framework can generalize across different backbone models in a plug-and-play manner.

\begin{table}[tb]
\centering
\resizebox{0.9\linewidth}{!}{ 
\begin{tabular}{l|c}
\toprule
\textbf{Model} & \textbf{Accuracy (EX\%)} \\
\midrule
Deepseek-V3 & 8.70 \\
GPT-4o & 10.17 \\
GPT-4.1 & 13.57 \\ 
Gemini-2.5-Pro & 10.17 \\
Claude-3.7-Sonnet & 13.99 \\
Claude-4.0-Sonnet & 11.51 \\
Seed-Coder-8B-Reasoning & 4.16 \\
DeepSeek-R1 & 9.98 \\
\textbf{o4-mini} & \textbf{14.96} \\
o3-mini & 9.11 \\ 
Qwen3-235B-A22B & 10.52 \\
Qwen3-Coder-480B-A35B-Instruct & 9.42 \\
\midrule
\textbf{+ IESR (Qwen2.5-Coder-7B)} & \textbf{21.66 (↑ 6.70)} \\
\textbf{+ IESR (XiYanSQL-QwenCoder-7B-2504)} & \textbf{24.28 (↑ 9.32)} \\
\textbf{+ IESR (OmniSQL-7B)} & \textbf{22.41 (↑ 7.45)} \\    
\textbf{+ IESR (Seed-Coder-8B)} & \textbf{22.11 (↑ 7.15)} \\      
\textbf{+ IESR (Qwen2.5-7B-Instruct)} & \textbf{16.69 (↑ 1.73)} \\                    
\textbf{+ IESR (Qwen3-8B)} & \textbf{18.23 (↑ 3.27)} \\                     
\bottomrule
\end{tabular}
}
\caption{Comparison with Baseline LLMs on the LogicCat dataset. Comparing with o4-mini, \textbf{EX} shows Execution Accuracy.}
\label{tab:llmcompare}
\end{table}

\section{Conclusion}
We propose IESR, a MCTS based framework that integrates information understanding and schema linking for complex Text-to-SQL reasoning. Our method achieves strong reasoning-oriented performance with medium backbones and exposes a controllable accuracy-cost trade-off through the rollout budget, as validated by comprehensive ablation studies.

\section*{Limitations}
While IESR demonstrates strong performance on complex reasoning–oriented Text-to-SQL benchmarks, it also has limitations. First, the framework relies on the quality of early semantic hypothesis extraction; errors in entity, unit, or formula identification may propagate into subsequent schema compression and search. Second, although MCTS enables structured exploration, it introduces additional inference cost compared to single-pass generation, and the search budget must be carefully tuned to avoid diminishing returns from noisy trajectories. Third, our reward design is execution- and consistency-based, which provides robust supervision without annotations but offers limited insight into intermediate reasoning errors. Moreover, our evaluation is conducted under controlled benchmark settings, and extending the framework to real-world databases with evolving schema or noisy execution environments remains an open challenge. Finally, we briefly discuss potential risks such as erroneous SQL execution and misinterpretation of complex queries, which could impact downstream decision-making if used without human oversight. We stress that automated Text-to-SQL systems should be deployed with appropriate validation in real-world scenarios.  

\section*{Acknowledgments}
This work was supported by the Key Program of the National Natural Science Foundation of China (Grant No. U23A20316), the General Program of the Natural Science Foundation of Henan Province (Grant No. 252300421877), the Henan Provincial Science and Technology Research Project (Grant No. 262102211057), and the Open Project of the Sichuan Provincial Key Laboratory of Philosophy and Social Science for Language Intelligence in Special Education (Grant No. YYZN-2026-1).


\bibliography{custom}

@article{Wei_Wang_Schuurmans_Bosma_Chi_Le_Zhou,  
 title={Chain of Thought Prompting Elicits Reasoning in Large Language Models}, 
 author={Wei, Jason and Wang, Xuezhi and Schuurmans, Dale and Bosma, Maarten and Chi, Ed and Le, Quoc and Zhou, Denny}, 
 language={en-US} 
 }

@misc{wang2023planandsolvepromptingimprovingzeroshot,
      title={Plan-and-Solve Prompting: Improving Zero-Shot Chain-of-Thought Reasoning by Large Language Models}, 
      author={Lei Wang and Wanyu Xu and Yihuai Lan and Zhiqiang Hu and Yunshi Lan and Roy Ka-Wei Lee and Ee-Peng Lim},
      year={2023},
      eprint={2305.04091},
      archivePrefix={arXiv},
      primaryClass={cs.CL},
      url={https://arxiv.org/abs/2305.04091}, 
}

@inproceedings{
shinn2023reflexion,
title={Reflexion: language agents with verbal reinforcement learning},
author={Noah Shinn and Federico Cassano and Ashwin Gopinath and Karthik R Narasimhan and Shunyu Yao},
booktitle={Thirty-seventh Conference on Neural Information Processing Systems},
year={2023},
url={https://openreview.net/forum?id=vAElhFcKW6}
}

@inproceedings{
lei2025spider,
title={Spider 2.0: Evaluating Language Models on Real-World Enterprise Text-to-{SQL} Workflows},
author={Fangyu Lei and Jixuan Chen and Yuxiao Ye and Ruisheng Cao and Dongchan Shin and Hongjin SU and ZHAOQING SUO and Hongcheng Gao and Wenjing Hu and Pengcheng Yin and Victor Zhong and Caiming Xiong and Ruoxi Sun and Qian Liu and Sida Wang and Tao Yu},
booktitle={The Thirteenth International Conference on Learning Representations},
year={2025},
url={https://openreview.net/forum?id=XmProj9cPs}
}

@misc{xu2025mtirsqlmultiturntoolintegratedreasoning,
      title={MTIR-SQL: Multi-turn Tool-Integrated Reasoning Reinforcement Learning for Text-to-SQL}, 
      author={Zekun Xu and Siyu Xia and Chuhuai Yue and Jiajun Chai and Mingxue Tian and Xiaohan Wang and Wei Lin and Haoxuan Li and Guojun Yin},
      year={2025},
      eprint={2510.25510},
      archivePrefix={arXiv},
      primaryClass={cs.AI},
      url={https://arxiv.org/abs/2510.25510}, 
}

@misc{yuan2025mctssqllightweightllmsmaster,
      title={MCTS-SQL: Light-Weight LLMs can Master the Text-to-SQL through Monte Carlo Tree Search}, 
      author={Shuozhi Yuan and Limin Chen and Miaomiao Yuan and Zhao Jin},
      year={2025},
      eprint={2501.16607},
      archivePrefix={arXiv},
      primaryClass={cs.DB},
      url={https://arxiv.org/abs/2501.16607}, 
}

@inproceedings{weng-etal-2025-graph,
    title = "Graph-Reward-{SQL}: Execution-Free Reinforcement Learning for Text-to-{SQL} via Graph Matching and Stepwise Reward",
    author = "Weng, Han  and
      Wu, Puzhen  and
      Longjie, Cui  and
      Zhan, Yi  and
      Liu, Boyi  and
      Song, Yuanfeng  and
      Zeng, Dun  and
      Yang, Yingxiang  and
      Zhang, Qianru  and
      Huang, Dong  and
      Yin, Xiaoming  and
      Sun, Yang  and
      Chen, Xing",
    editor = "Christodoulopoulos, Christos  and
      Chakraborty, Tanmoy  and
      Rose, Carolyn  and
      Peng, Violet",
    booktitle = "Findings of the Association for Computational Linguistics: EMNLP 2025",
    month = nov,
    year = "2025",
    address = "Suzhou, China",
    publisher = "Association for Computational Linguistics",
    url = "https://aclanthology.org/2025.findings-emnlp.694/",
    doi = "10.18653/v1/2025.findings-emnlp.694",
    pages = "12917--12943",
    ISBN = "979-8-89176-335-7"
}

@misc{yao2026arctictext2sqlr1simplerewardsstrong,
      title={Arctic-Text2SQL-R1: Simple Rewards, Strong Reasoning in Text-to-SQL}, 
      author={Zhewei Yao and Guoheng Sun and Lukasz Borchmann and Gaurav Nuti and Zheyu Shen and Minghang Deng and Bohan Zhai and Hao Zhang and Ang Li and Yuxiong He},
      year={2026},
      eprint={2505.20315},
      archivePrefix={arXiv},
      primaryClass={cs.CL},
      url={https://arxiv.org/abs/2505.20315}, 
}

@inproceedings{toteja-etal-2025-context,
    title = "In-Context Reinforcement Learning with Retrieval-Augmented Generation for Text-to-{SQL}",
    author = "Toteja, Rishit  and
      Sarkar, Arindam  and
      Comar, Prakash Mandayam",
    editor = "Rambow, Owen  and
      Wanner, Leo  and
      Apidianaki, Marianna  and
      Al-Khalifa, Hend  and
      Eugenio, Barbara Di  and
      Schockaert, Steven",
    booktitle = "Proceedings of the 31st International Conference on Computational Linguistics",
    month = jan,
    year = "2025",
    address = "Abu Dhabi, UAE",
    publisher = "Association for Computational Linguistics",
    url = "https://aclanthology.org/2025.coling-main.692/",
    pages = "10390--10397"
}

@misc{wang2025agentarscalesqladvancingtexttosqlorchestrated,
      title={Agentar-Scale-SQL: Advancing Text-to-SQL through Orchestrated Test-Time Scaling}, 
      author={Pengfei Wang and Baolin Sun and Xuemei Dong and Yaxun Dai and Hongwei Yuan and Mengdie Chu and Yingqi Gao and Xiang Qi and Peng Zhang and Ying Yan},
      year={2025},
      eprint={2509.24403},
      archivePrefix={arXiv},
      primaryClass={cs.CL},
      url={https://arxiv.org/abs/2509.24403}, 
}

@misc{xia2024agentlessdemystifyingllmbasedsoftware,
      title={Agentless: Demystifying LLM-based Software Engineering Agents}, 
      author={Chunqiu Steven Xia and Yinlin Deng and Soren Dunn and Lingming Zhang},
      year={2024},
      eprint={2407.01489},
      archivePrefix={arXiv},
      primaryClass={cs.SE},
      url={https://arxiv.org/abs/2407.01489}, 
}

@misc{talaei2024chesscontextualharnessingefficient,
      title={CHESS: Contextual Harnessing for Efficient SQL Synthesis}, 
      author={Shayan Talaei and Mohammadreza Pourreza and Yu-Chen Chang and Azalia Mirhoseini and Amin Saberi},
      year={2024},
      eprint={2405.16755},
      archivePrefix={arXiv},
      primaryClass={cs.LG},
      url={https://arxiv.org/abs/2405.16755}, 
}

@inproceedings{
pourreza2025chasesql,
title={{CHASE}-{SQL}: Multi-Path Reasoning and Preference Optimized Candidate Selection in Text-to-{SQL}},
author={Mohammadreza Pourreza and Hailong Li and Ruoxi Sun and Yeounoh Chung and Shayan Talaei and Gaurav Tarlok Kakkar and Yu Gan and Amin Saberi and Fatma Ozcan and Sercan O Arik},
booktitle={The Thirteenth International Conference on Learning Representations},
year={2025},
url={https://openreview.net/forum?id=CvGqMD5OtX}
}

@article{Zelle_Mooney_1996,   
title={Learning to parse database queries using inductive logic programming},  
journal={National Conference on Artificial Intelligence,National Conference on Artificial Intelligence},  
author={Zelle, JohnM. and Mooney, RaymondJ.},  
year={1996},  
month={Aug},  
language={en-US}  
}

@misc{yu2019spiderlargescalehumanlabeleddataset,
      title={Spider: A Large-Scale Human-Labeled Dataset for Complex and Cross-Domain Semantic Parsing and Text-to-SQL Task}, 
      author={Tao Yu and Rui Zhang and Kai Yang and Michihiro Yasunaga and Dongxu Wang and Zifan Li and James Ma and Irene Li and Qingning Yao and Shanelle Roman and Zilin Zhang and Dragomir Radev},
      year={2019},
      eprint={1809.08887},
      archivePrefix={arXiv},
      primaryClass={cs.CL},
      url={https://arxiv.org/abs/1809.08887}, 
}

@misc{wretblad2024understandingeffectsnoisetexttosql,
      title={Understanding the Effects of Noise in Text-to-SQL: An Examination of the BIRD-Bench Benchmark}, 
      author={Niklas Wretblad and Fredrik Gordh Riseby and Rahul Biswas and Amin Ahmadi and Oskar Holmström},
      year={2024},
      eprint={2402.12243},
      archivePrefix={arXiv},
      primaryClass={cs.CL},
      url={https://arxiv.org/abs/2402.12243}, 
}

@article{Li_2024,
   title={The Dawn of Natural Language to SQL: Are We Fully Ready?},
   volume={17},
   ISSN={2150-8097},
   url={http://dx.doi.org/10.14778/3681954.3682003},
   DOI={10.14778/3681954.3682003},
   number={11},
   journal={Proceedings of the VLDB Endowment},
   publisher={Association for Computing Machinery (ACM)},
   author={Li, Boyan and Luo, Yuyu and Chai, Chengliang and Li, Guoliang and Tang, Nan},
   year={2024},
   month=jul, pages={3318–3331} }

@misc{cao2024rslsqlrobustschemalinking,
      title={RSL-SQL: Robust Schema Linking in Text-to-SQL Generation}, 
      author={Zhenbiao Cao and Yuanlei Zheng and Zhihao Fan and Xiaojin Zhang and Wei Chen and Xiang Bai},
      year={2024},
      eprint={2411.00073},
      archivePrefix={arXiv},
      primaryClass={cs.CL},
      url={https://arxiv.org/abs/2411.00073}, 
}

@inproceedings{
li2025alphasql,
title={Alpha-{SQL}: Zero-Shot Text-to-{SQL} using Monte Carlo Tree Search},
author={Boyan Li and Jiayi Zhang and Ju Fan and Yanwei Xu and Chong Chen and Nan Tang and Yuyu Luo},
booktitle={Forty-second International Conference on Machine Learning},
year={2025},
url={https://openreview.net/forum?id=kGg1ndttmI}
}

@misc{lyu2025sqlo1selfrewardheuristicdynamic,
      title={SQL-o1: A Self-Reward Heuristic Dynamic Search Method for Text-to-SQL}, 
      author={Shuai Lyu and Haoran Luo and Ripeng Li and Zhonghong Ou and Jiangfeng Sun and Yang Qin and Xiaoran Shang and Meina Song and Yifan Zhu},
      year={2025},
      eprint={2502.11741},
      archivePrefix={arXiv},
      primaryClass={cs.DB},
      url={https://arxiv.org/abs/2502.11741}, 
}

@misc{chen2024alphamathzeroprocesssupervision,
      title={AlphaMath Almost Zero: Process Supervision without Process}, 
      author={Guoxin Chen and Minpeng Liao and Chengxi Li and Kai Fan},
      year={2024},
      eprint={2405.03553},
      archivePrefix={arXiv},
      primaryClass={cs.CL},
      url={https://arxiv.org/abs/2405.03553}, 
}

@misc{liu2025logiccatchainofthoughttexttosqlbenchmark,
      title={LogicCat: A Chain-of-Thought Text-to-SQL Benchmark for Multi-Domain Reasoning Challenges}, 
      author={Tao Liu and Hongying Zan and Yifan Li and Dixuan Zhang and Lulu Kong and Haixin Liu and Jiaming Hou and Aoze Zheng and Rui Li and Yiming Qiao and Zewei Luo and Qi Wang and Zhiqiang Zhang and Jiaxi Li and Supeng Liu and Kunli Zhang and Min Peng},
      year={2025},
      eprint={2505.18744},
      archivePrefix={arXiv},
      primaryClass={cs.CL},
      url={https://arxiv.org/abs/2505.18744}, 
}

@inproceedings{zheng-etal-2024-archer,
    title = "Archer: A Human-Labeled Text-to-{SQL} Dataset with Arithmetic, Commonsense and Hypothetical Reasoning",
    author = "Zheng, Danna  and
      Lapata, Mirella  and
      Pan, Jeff",
    editor = "Graham, Yvette  and
      Purver, Matthew",
    booktitle = "Proceedings of the 18th Conference of the European Chapter of the Association for Computational Linguistics (Volume 1: Long Papers)",
    month = mar,
    year = "2024",
    address = "St. Julian{'}s, Malta",
    publisher = "Association for Computational Linguistics",
    url = "https://aclanthology.org/2024.eacl-long.6/",
    doi = "10.18653/v1/2024.eacl-long.6",
    pages = "94--111"
}

@article{XiYanSQL,
      title={XiYan-SQL: A Novel Multi-Generator Framework For Text-to-SQL}, 
      author={Yifu Liu and Yin Zhu and Yingqi Gao and Zhiling Luo and Xiaoxia Li and Xiaorong Shi and Yuntao Hong and Jinyang Gao and Yu Li and Bolin Ding and Jingren Zhou},
      year={2025},
      eprint={2507.04701},
      archivePrefix={arXiv},
      primaryClass={cs.CL},
      url={https://arxiv.org/abs/2507.04701}, 
}

@misc{hui2024qwen25codertechnicalreport,
      title={Qwen2.5-Coder Technical Report}, 
      author={Binyuan Hui and Jian Yang and Zeyu Cui and Jiaxi Yang and Dayiheng Liu and Lei Zhang and Tianyu Liu and Jiajun Zhang and Bowen Yu and Keming Lu and Kai Dang and Yang Fan and Yichang Zhang and An Yang and Rui Men and Fei Huang and Bo Zheng and Yibo Miao and Shanghaoran Quan and Yunlong Feng and Xingzhang Ren and Xuancheng Ren and Jingren Zhou and Junyang Lin},
      year={2024},
      eprint={2409.12186},
      archivePrefix={arXiv},
      primaryClass={cs.CL},
      url={https://arxiv.org/abs/2409.12186}, 
}

@misc{gao2025previewxiyansqlmultigeneratorensemble,
      title={A Preview of XiYan-SQL: A Multi-Generator Ensemble Framework for Text-to-SQL}, 
      author={Yingqi Gao and Yifu Liu and Xiaoxia Li and Xiaorong Shi and Yin Zhu and Yiming Wang and Shiqi Li and Wei Li and Yuntao Hong and Zhiling Luo and Jinyang Gao and Liyu Mou and Yu Li},
      year={2025},
      eprint={2411.08599},
      archivePrefix={arXiv},
      primaryClass={cs.AI},
      url={https://arxiv.org/abs/2411.08599}, 
}

@misc{seed2025seedcoderletcodemodel,
      title={Seed-Coder: Let the Code Model Curate Data for Itself}, 
      author={ByteDance Seed and Yuyu Zhang and Jing Su and Yifan Sun and Chenguang Xi and Xia Xiao and Shen Zheng and Anxiang Zhang and Kaibo Liu and Daoguang Zan and Tao Sun and Jinhua Zhu and Shulin Xin and Dong Huang and Yetao Bai and Lixin Dong and Chao Li and Jianchong Chen and Hanzhi Zhou and Yifan Huang and Guanghan Ning and Xierui Song and Jiaze Chen and Siyao Liu and Kai Shen and Liang Xiang and Yonghui Wu},
      year={2025},
      eprint={2506.03524},
      archivePrefix={arXiv},
      primaryClass={cs.CL},
      url={https://arxiv.org/abs/2506.03524}, 
}

@misc{li2025omnisqlsynthesizinghighqualitytexttosql,
      title={OmniSQL: Synthesizing High-quality Text-to-SQL Data at Scale}, 
      author={Haoyang Li and Shang Wu and Xiaokang Zhang and Xinmei Huang and Jing Zhang and Fuxin Jiang and Shuai Wang and Tieying Zhang and Jianjun Chen and Rui Shi and Hong Chen and Cuiping Li},
      year={2025},
      eprint={2503.02240},
      archivePrefix={arXiv},
      primaryClass={cs.CL},
      url={https://arxiv.org/abs/2503.02240}, 
}

@misc{yang2025qwen3technicalreport,
      title={Qwen3 Technical Report}, 
      author={An Yang and Anfeng Li and Baosong Yang and Beichen Zhang and Binyuan Hui and Bo Zheng and Bowen Yu and Chang Gao and Chengen Huang and Chenxu Lv and Chujie Zheng and Dayiheng Liu and Fan Zhou and Fei Huang and Feng Hu and Hao Ge and Haoran Wei and Huan Lin and Jialong Tang and Jian Yang and Jianhong Tu and Jianwei Zhang and Jianxin Yang and Jiaxi Yang and Jing Zhou and Jingren Zhou and Junyang Lin and Kai Dang and Keqin Bao and Kexin Yang and Le Yu and Lianghao Deng and Mei Li and Mingfeng Xue and Mingze Li and Pei Zhang and Peng Wang and Qin Zhu and Rui Men and Ruize Gao and Shixuan Liu and Shuang Luo and Tianhao Li and Tianyi Tang and Wenbiao Yin and Xingzhang Ren and Xinyu Wang and Xinyu Zhang and Xuancheng Ren and Yang Fan and Yang Su and Yichang Zhang and Yinger Zhang and Yu Wan and Yuqiong Liu and Zekun Wang and Zeyu Cui and Zhenru Zhang and Zhipeng Zhou and Zihan Qiu},
      year={2025},
      eprint={2505.09388},
      archivePrefix={arXiv},
      primaryClass={cs.CL},
      url={https://arxiv.org/abs/2505.09388}, 
}

@inproceedings{
pourreza2023dinsql,
title={{DIN}-{SQL}: Decomposed In-Context Learning of Text-to-{SQL} with Self-Correction},
author={Mohammadreza Pourreza and Davood Rafiei},
booktitle={Thirty-seventh Conference on Neural Information Processing Systems},
year={2023},
url={https://openreview.net/forum?id=p53QDxSIc5}
}

@misc{gao2023texttosqlempoweredlargelanguage,
      title={Text-to-SQL Empowered by Large Language Models: A Benchmark Evaluation}, 
      author={Dawei Gao and Haibin Wang and Yaliang Li and Xiuyu Sun and Yichen Qian and Bolin Ding and Jingren Zhou},
      year={2023},
      eprint={2308.15363},
      archivePrefix={arXiv},
      primaryClass={cs.DB},
      url={https://arxiv.org/abs/2308.15363}, 
}

@misc{wang2025linkalignscalableschemalinking,
      title={LinkAlign: Scalable Schema Linking for Real-World Large-Scale Multi-Database Text-to-SQL}, 
      author={Yihan Wang and Peiyu Liu},
      year={2025},
      eprint={2503.18596},
      archivePrefix={arXiv},
      primaryClass={cs.CL},
      url={https://arxiv.org/abs/2503.18596}, 
}

@inproceedings{pourreza-rafiei-2024-dts,
    title = "{DTS}-{SQL}: Decomposed Text-to-{SQL} with Small Large Language Models",
    author = "Pourreza, Mohammadreza  and
      Rafiei, Davood",
    editor = "Al-Onaizan, Yaser  and
      Bansal, Mohit  and
      Chen, Yun-Nung",
    booktitle = "Findings of the Association for Computational Linguistics: EMNLP 2024",
    month = nov,
    year = "2024",
    address = "Miami, Florida, USA",
    publisher = "Association for Computational Linguistics",
    url = "https://aclanthology.org/2024.findings-emnlp.481/",
    doi = "10.18653/v1/2024.findings-emnlp.481",
    pages = "8212--8220"
}

@misc{deng2025reforcetexttosqlagentselfrefinement,
      title={ReFoRCE: A Text-to-SQL Agent with Self-Refinement, Consensus Enforcement, and Column Exploration}, 
      author={Minghang Deng and Ashwin Ramachandran and Canwen Xu and Lanxiang Hu and Zhewei Yao and Anupam Datta and Hao Zhang},
      year={2025},
      eprint={2502.00675},
      archivePrefix={arXiv},
      primaryClass={cs.CL},
      url={https://arxiv.org/abs/2502.00675}, 
}

@misc{yao2023reactsynergizingreasoningacting,
      title={ReAct: Synergizing Reasoning and Acting in Language Models}, 
      author={Shunyu Yao and Jeffrey Zhao and Dian Yu and Nan Du and Izhak Shafran and Karthik Narasimhan and Yuan Cao},
      year={2023},
      eprint={2210.03629},
      archivePrefix={arXiv},
      primaryClass={cs.CL},
      url={https://arxiv.org/abs/2210.03629}, 
}

@misc{song2025joltsqljointlosstuning,
      title={JOLT-SQL: Joint Loss Tuning of Text-to-SQL with Confusion-aware Noisy Schema Sampling}, 
      author={Jinwang Song and Hongying Zan and Kunli Zhang and Lingling Mu and Yingjie Han and Haobo Hua and Min Peng},
      year={2025},
      eprint={2505.14305},
      archivePrefix={arXiv},
      primaryClass={cs.CL},
      url={https://arxiv.org/abs/2505.14305}, 
}

@inproceedings{
qi2025mutual,
title={Mutual Reasoning Makes Smaller {LLM}s Stronger Problem-Solver},
author={Zhenting Qi and Mingyuan MA and Jiahang Xu and Li Lyna Zhang and Fan Yang and Mao Yang},
booktitle={The Thirteenth International Conference on Learning Representations},
year={2025},
url={https://openreview.net/forum?id=6aHUmotXaw}
}

@misc{mao2025steinersqlgraphguidedmathematicalreasoning,
      title={SteinerSQL: Graph-Guided Mathematical Reasoning for Text-to-SQL Generation}, 
      author={Xutao Mao and Tao Liu and Hongying Zan},
      year={2025},
      eprint={2509.19623},
      archivePrefix={arXiv},
      primaryClass={cs.AI},
      url={https://arxiv.org/abs/2509.19623}, 
}

@misc{li2025deepeyesqlsoftwareengineeringinspiredtexttosqlframework,
      title={DeepEye-SQL: A Software-Engineering-Inspired Text-to-SQL Framework}, 
      author={Boyan Li and Chong Chen and Zhujun Xue and Yinan Mei and Yuyu Luo},
      year={2025},
      eprint={2510.17586},
      archivePrefix={arXiv},
      primaryClass={cs.DB},
      url={https://arxiv.org/abs/2510.17586}, 
}

@article{yao2025arctic,
  title={Arctic-Text2SQL-R1: Simple Rewards, Strong Reasoning in Text-to-SQL},
  author={Yao, Zhewei and others},
  year={2025},
  journal={arXiv preprint},
}

@inproceedings{10.1609/aaai.v39i1.31979,
author = {Cen, Jipeng and Liu, Jiaxin and Li, Zhixu and Wang, Jingjing},
title = {SQLFixAgent: towards semantic-accurate text-to-SQL parsing via consistency-enhanced multi-agent collaboration},
year = {2025},
isbn = {978-1-57735-897-8},
publisher = {AAAI Press},
url = {https://doi.org/10.1609/aaai.v39i1.31979},
doi = {10.1609/aaai.v39i1.31979},
booktitle = {Proceedings of the Thirty-Ninth AAAI Conference on Artificial Intelligence and Thirty-Seventh Conference on Innovative Applications of Artificial Intelligence and Fifteenth Symposium on Educational Advances in Artificial Intelligence},
articleno = {6},
numpages = {9},
series = {AAAI'25/IAAI'25/EAAI'25}
}

@misc{wang2025autolinkautonomousschemaexploration,
      title={AutoLink: Autonomous Schema Exploration and Expansion for Scalable Schema Linking in Text-to-SQL at Scale}, 
      author={Ziyang Wang and Yuanlei Zheng and Zhenbiao Cao and Xiaojin Zhang and Zhongyu Wei and Pei Fu and Zhenbo Luo and Wei Chen and Xiang Bai},
      year={2025},
      eprint={2511.17190},
      archivePrefix={arXiv},
      primaryClass={cs.CL},
      url={https://arxiv.org/abs/2511.17190}, 
}

@inproceedings{10.1145/3772318.3791724,
author = {Zhou, Zihan and Liu, Yinan and Xie, Yuyang and Wang, Bin and Yang, Xiaochun and Feng, Zezheng},
title = {DiagLink: A Dual-User Diagnostic Assistance System by Synergizing Experts with LLMs and Knowledge Graphs},
year = {2026},
isbn = {9798400722783},
publisher = {Association for Computing Machinery},
address = {New York, NY, USA},
url = {https://doi.org/10.1145/3772318.3791724},
doi = {10.1145/3772318.3791724},
booktitle = {Proceedings of the 2026 CHI Conference on Human Factors in Computing Systems},
articleno = {142},
numpages = {28},
}

@misc{liu2026enhancingllmspredictivepolitical,
      title={Enhancing LLMs in Predictive Political QA with Semi-Structured Data}, 
      author={Yinan Liu and Zihan Zhou and Zichun Jin and Xinyu Wang and Bin Wang and Xiaochun Yang},
      year={2026},
      eprint={2608.21218},
      archivePrefix={arXiv},
      primaryClass={cs.AI},
      url={https://arxiv.org/abs/2608.21218}, 
}

@misc{jin2026adaptingknowledgegraphsbehavior,
      title={Adapting Knowledge Graphs for Behavior Denoising in Sequential Recommendation}, 
      author={Zichun Jin and Zihan Zhou and Yinan Liu and Bin Wang and Xiaochun Yang},
      year={2026},
      eprint={2608.21243},
      archivePrefix={arXiv},
      primaryClass={cs.IR},
      url={https://arxiv.org/abs/2608.21243}, 
}

@inproceedings{feng-etal-2026-trust,
    title = "Trust Within? Seek Beyond? Knowledge Boundary Aware Policy Optimization for Agentic Search",
    author = "Feng, Tao  and
      Jiang, Xinke  and
      Hu, Xinyan  and
      Zhang, Yonggang  and
      Tao, Zhen  and
      Zhang, Wentao  and
      Liu, Boyang  and
      Jiang, Wenhao  and
      Wu, Chao",
    editor = "Liakata, Maria  and
      Moreira, Viviane P.  and
      Zhang, Jiajun  and
      Jurgens, David",
    booktitle = "Proceedings of the 64th Annual Meeting of the {A}ssociation for {C}omputational {L}inguistics (Volume 1: Long Papers)",
    month = jul,
    year = "2026",
    address = "San Diego, California, United States",
    publisher = "Association for Computational Linguistics",
    url = "https://aclanthology.org/2026.acl-long.1276/",
    doi = "10.18653/v1/2026.acl-long.1276",
    pages = "27664--27682",
    ISBN = "979-8-89176-390-6"
}

@misc{feng2026kbsdknowledgeboundaryaware,
      title={KbSD: Knowledge Boundary aware Self-Distillation for Behavioral Calibration in Agentic Search}, 
      author={Tao Feng and Xinke Jiang and Chao Wu},
      year={2026},
      eprint={2606.29863},
      archivePrefix={arXiv},
      primaryClass={cs.CL},
      url={https://arxiv.org/abs/2606.29863}, 
}

@misc{liu2026serlsqlselectivehindsightdistillation,
      title={SERL-SQL: Selective Hindsight Distillation for Text-to-SQL Reinforcement Agentic Learning}, 
      author={Tao Liu and Tao Feng and Xiangheng Li and Jinwang Song and Yifan Li and Xiaoqing Cheng and Dixuan Zhang and Siquan Li and Lin Lan and Hongying Zan and Kunli Zhang and Chao Wu},
      year={2026},
      eprint={2608.00485},
      archivePrefix={arXiv},
      primaryClass={cs.CL},
      url={https://arxiv.org/abs/2608.00485}, 
}

@inproceedings{liu2025knowledge,
  title={Knowledge-Enhanced and Event-Rule Guided Framework for Fine-Grained Argument Mining in Chinese Essays},
  author={Liu, Tao and Li, Yunlong and Li, Yifan and Yu, Bohan and Zheng, Aoze and Zhang, Kunli and Zan, Hongying and Peng, Ming},
  booktitle={CCF International Conference on Natural Language Processing and Chinese Computing},
  pages={347--357},
  year={2025},
  organization={Springer}
}

\appendix

\section{Experimental Analysis}
\label{sec:general_benchmark_results}
\subsection{Comparison on General datasets}
Table~\ref{tab:general_benchmark_results} reports results on BIRD-Dev and Spider. These benchmarks are included to characterize the transfer behavior of IESR beyond reasoning-intensive queries. The results show that explicit multi-path reasoning does not consistently improve standard Text-to-SQL benchmarks, where specialized schema-linking and candidate-ranking methods remain highly competitive.

\begin{table*}[t]
\centering
\renewcommand{\arraystretch}{0.92}
\setlength{\tabcolsep}{0.8mm}
\footnotesize
\begin{tabular}{l|cccc|cccc}
\toprule
\textbf{Method}
\textbf{Method} & \textbf{Qwen2.5-Coder} & \textbf{XiYanSQL} & \multicolumn{1}{c|}{\textbf{OmniSQL}} & \textbf{Seed-Coder} & \textbf{Qwen2.5-Coder} & \textbf{XiYanSQL} & \multicolumn{1}{c|}{\textbf{OmniSQL}} & \textbf{Seed-Coder} \\
\midrule
\textbf{Size}
& \multicolumn{3}{c|}{7B} & 8B
& \multicolumn{3}{c|}{7B} & 8B \\
\midrule
\multicolumn{1}{c|}{}
& \multicolumn{4}{c|}{\textbf{BIRD-Dev EX(\%)}}
& \multicolumn{4}{c}{\textbf{Spider EX(\%)}} \\
\midrule
Cot-SQL      & 31.22 & 33.58 & 37.58 & 30.16 & 65.32 & 69.12 & 68.32 & 66.23 \\
DIN-SQL      & 51.78 & 52.45 & 51.23 & 49.88 & 77.19 & 79.30 & 76.80 & 78.01 \\
DAIL-SQL     & 52.12 & 51.16 & 52.23 & 50.66 & 80.31 & 82.56 & 81.56 & 82.98 \\
DTS-SQL      & 58.32 & 60.78 & 61.56 & 58.18 & 84.12 & 85.09 & 84.99 & 83.21 \\
CHESS-SQL    & 56.80 & 60.13 & 57.96 & 57.22 & 80.31 & 83.30 & 77.02 & 81.31 \\
Chase-SQL    & 58.99 & 61.68 & 56.23 & 62.17 & \textbf{85.11} & 85.12 & 81.42 & 80.15 \\
Deepeye-SQL  & 63.13 & 67.27 & \textbf{66.56} & \textbf{65.21} & 83.38 & \textbf{86.89} & \textbf{85.14} & 83.78 \\
LinkAlign    & 58.90 & 61.32 & 60.56 & 54.78 & 80.11 & 85.89 & 82.32 & 82.12 \\
SQL-O1       & 66.70 & 67.21 & 61.39 & 64.67 & 85.10 & 86.54 & 81.12 & \textbf{85.25} \\
Alpha-SQL    & \textbf{66.80} & \textbf{67.92} & 62.32 & 65.13 & 84.00 & 84.93 & 79.34 & 84.37 \\
\textbf{IESR}
& 54.80 & 62.22 & 56.22 & 58.96
& 83.45 & 83.36 & 80.02 & 79.22 \\
\bottomrule
\end{tabular}
\caption{Results on conventional Text-to-SQL benchmarks. These results are reported for completeness and clarify the scope of IESR. IESR is primarily designed for reasoning-intensive queries, while specialized Text-to-SQL systems remain stronger on standard schema-grounding benchmarks.}
\label{tab:general_benchmark_results}
\end{table*}

\label{sec:compariAnalysis}

\subsection{Detailed  $N_\text{rollout}$ Analysis}
\label{sec:rolloutAnalysis}
As shown in Figure~\ref{fig:rollouts}, disabling schema compression leads to a consistent performance drop across all backbones, highlighting the importance of schema understanding in constraining the search space for Text-to-SQL reasoning. Without compression, the reasoning agent must explore a larger set of tables and columns, which increases the likelihood of semantically mis-grounded yet executable SQL trajectories and reduces the stability of downstream aggregation.

We further observe that increasing $N_{\text{rollout}}$ generally improves performance up to $N_{\text{rollout}}=32$, where all models achieve their best results. This improvement can be attributed to enhanced exploration coverage and more reliable consensus estimation: a larger rollout budget increases the probability of sampling high-quality reasoning trajectories while reducing variance through majority voting.

However, performance gains saturate and may slightly decline beyond this point. We attribute this behavior to noise amplification effects: additional rollouts increasingly sample low-probability or semantically inconsistent trajectories, which can dilute the majority signal and introduce competition between incorrect but internally consistent modes. As a result, excessive rollouts do not provide independent evidence but may instead reinforce systematic biases of the backbone model.

Overall, these results indicate that a moderate rollout budget strikes the best balance between exploration and aggregation reliability. In practice, $N_{\text{rollout}}=32$ offers a strong trade-off between robustness and noise, while larger budgets yield diminishing or even negative returns.

\begin{table}[htbp]
\centering
\small
\renewcommand{\arraystretch}{0.92}
\setlength{\tabcolsep}{0.5mm}
\resizebox{\columnwidth}{!}{%
\begin{tabular}{lcccccccc}
\toprule
& \multicolumn{2}{c}{\textbf{Qwen2.5-Coder}} 
& \multicolumn{2}{c}{\textbf{OmniSQL}} 
& \multicolumn{2}{c}{\textbf{Seed-Coder}} 
& \multicolumn{2}{c}{\textbf{XiYanSQL}} \\
\cmidrule(lr){2-3} \cmidrule(lr){4-5} \cmidrule(lr){6-7} \cmidrule(lr){8-9}
\textbf{Configuration} 
& \textbf{EX} & \textbf{Drop} 
& \textbf{EX} & \textbf{Drop} 
& \textbf{EX} & \textbf{Drop} 
& \textbf{EX} & \textbf{Drop} \\
\midrule
\textbf{SC@maj32}  
& \textbf{21.66} & -- 
& \textbf{24.28} & -- 
& \textbf{22.11} & -- 
& \textbf{22.41} & -- \\
\midrule
\multicolumn{9}{l}{\textit{Ablation $N_\text{rollout}$}} \\
SC@maj8  
& 20.68 & 0.98 
& 23.22 & 1.06 
& 21.18 & 0.93 
& 20.45 & 1.96 \\
SC@maj16 
& 20.40 & 1.26 
& 23.86 & 0.42 
& 21.45 & 0.66 
& 20.81 & 1.60 \\
SC@maj24 
& 20.91 & 0.75 
& 22.89 & 1.39 
& 21.85 & 0.26 
& 21.41 & 1.00 \\
SC@maj40 
& 21.12 & 0.54 
& 23.86 & 0.42 
& 21.45 & 0.66 
& 20.98 & 1.43 \\
\bottomrule
\end{tabular}%
}
\caption{Ablation study of $N_\text{rollout}$ across four backbone models on the LogicCat dataset. We report \textbf{EX} and the absolute performance drop (\%). \textbf{XiYanSQL} refers to XiYanSQL-QwenCoder-7B-2504; \textbf{Qwen2.5-Coder} to Qwen2.5-Coder-7B; \textbf{OmniSQL} to OmniSQL-7B; and \textbf{Seed-Coder} to Seed-Coder-8B.}
\label{tab:ablation_four_Nrollout}
\end{table}

\subsection{Perturbation Experiment Analysis}
\label{sec:PerturbationAnalysis}
We evaluated the robustness of IESR with a systematic perturbation study on 500 LogicCat questions with explicit JOIN intent. We construct three decay levels of increasing distractor complexity and compare IESR to standard prompting across four backbones, controlling for model capacity so that performance gaps reflect the framework itself.
\begin{itemize}
    \item \textbf{L1-Semantic:} Semantically related tables with misleading column names.
    \item \textbf{L2-Structural:} Similar table names with irrelevant columns or redundant JOIN paths.
    \item \textbf{L3-Hybrid Advanced:} Semantically relevant and structurally valid decays that violate specific constraints.
\end{itemize}

For each question, we create four database variants: \textit{DB\_Original}, \textit{DB\_Decay\_L1}, \textit{DB\_Decay\_L2}, and \textit{DB\_Decay\_L3}, testing both methods across all environments.

\begin{figure*}[ht]
    \centering
    
    \begin{minipage}{0.48\textwidth}
     \centering
     \includegraphics[width=\linewidth]{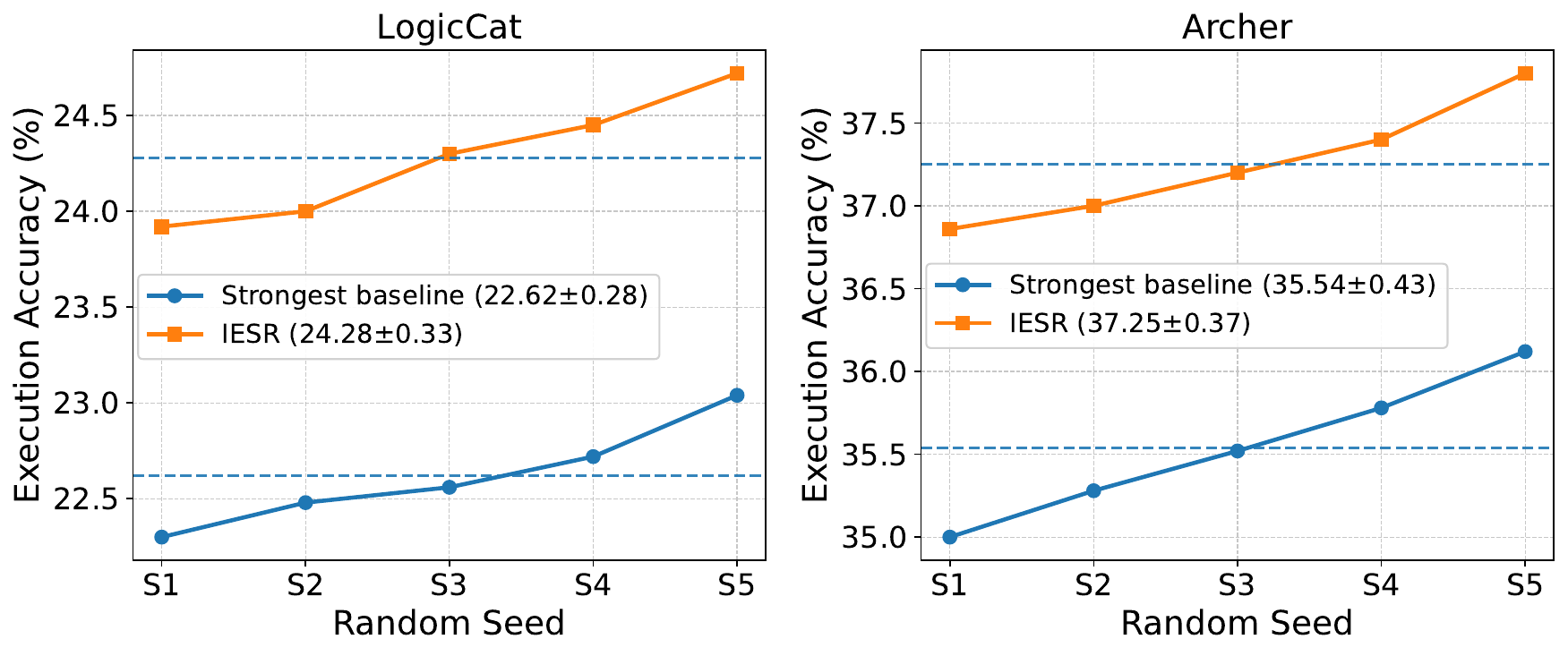}
     \caption{Execution accuracy across five random seeds on LogicCat and Archer.
    Solid lines show per-seed EX for the strongest baseline and IESR.
    Dashed horizontal lines indicate the mean EX, and shaded bands denote the 95\% confidence interval across seeds.}
         \label{fig:seed_stability}  
    \end{minipage}\hfill
    \begin{minipage}{0.48\textwidth}
        \centering
        \includegraphics[width=\linewidth]{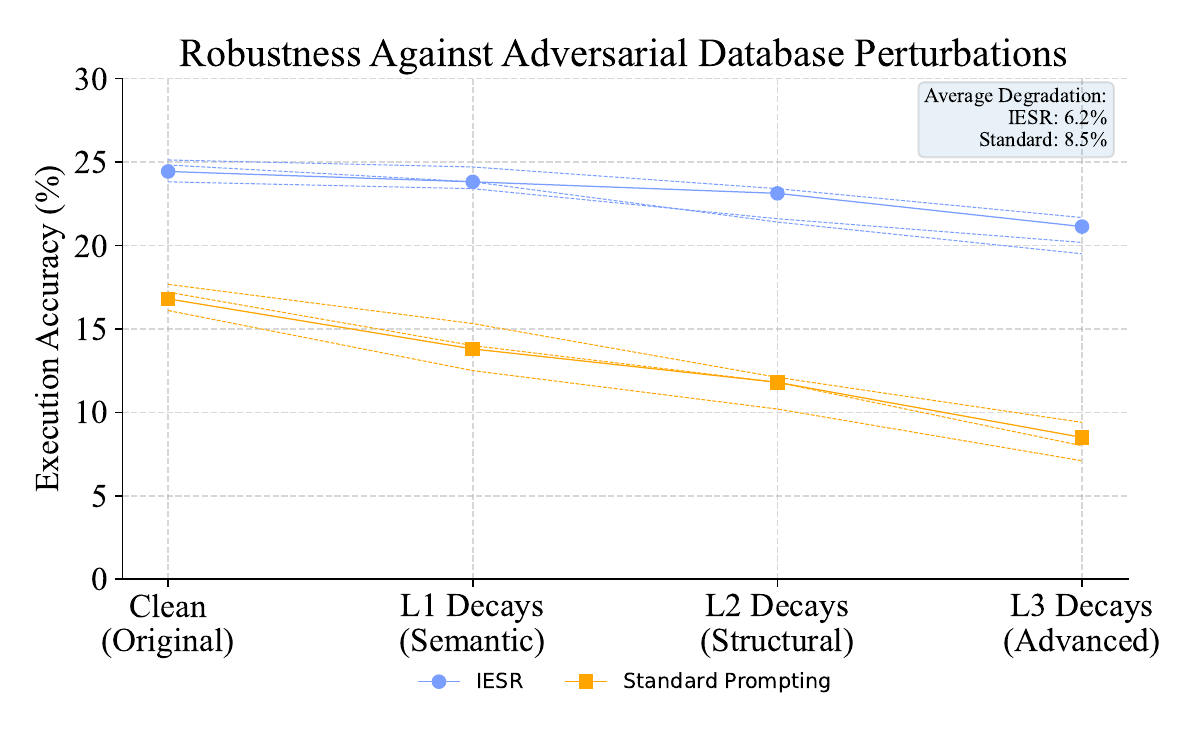}
        \caption{Robustness comparison between IESR and standard prompting across adversarial database environments. IESR maintains superior performance and exhibits more graceful degradation as decay sophistication increases from clean databases to advanced hybrid decays.}
        \label{fig:perturbation_robustness}
    \end{minipage}
\end{figure*}


Figure~\ref{fig:perturbation_robustness} presents the experimental results in all combinations of model-environment combinations, revealing the superior robustness of IESR against adversarial database perturbations~\cite{10.1145/3772318.3791724}. The results demonstrate IESR's consistent outperformance of standard prompting across all environments, with performance gaps that widen as adversarial complexity increases. In clean databases, IESR establishes a clear baseline advantage that expands progressively through L1 semantic decays and L2 structural decays and culminates in the largest gap under advanced hybrid decays L3~\cite{liu2026enhancingllmspredictivepolitical}.

IESR exhibits better resilience against adversarial perturbations, showing more graceful degradation compared to the steep decline in standard prompting. This robustness advantage stems from the framework's dual-component architecture: MCTS-based reasoning Configuration systematically breaks down complex queries while Trajectory Selection bridging ensures consistent relationship traversal, effectively avoiding the semantically plausible but logically incorrect decay paths that mislead standard approaches.

\subsection{Stability Analysis}
\label{sec:StabilityAnalysis}
Figure~\ref{fig:seed_stability} evaluates the robustness of IESR under different random seeds. We compare IESR with the strongest baseline on each dataset using the same backbone, decoding configuration, rollout budget, and execution backend. Across five independent runs, IESR consistently outperforms the strongest baseline on both LogicCat and Archer. On LogicCat, IESR achieves an average EX of 24.28\% with a standard deviation of 0.33, compared with 22.56\% and 0.35 for the strongest baseline. On Archer, IESR reaches 37.28\% with a standard deviation of 0.37, compared with 35.54\% and 0.43 for the strongest baseline. Although IESR shows comparable variance due to stochastic MCTS exploration and trajectory selection, its lower confidence bound remains above the upper confidence bound of the strongest baseline on both datasets. This indicates that the improvement is stable across random seeds and is unlikely to result from a favorable sampling instance.

\subsection{Efficiency Analysis}
\label{sec:efficiencyanalysis}

IESR introduces additional inference cost because it treats Text-to-SQL generation as a search problem rather than a single-pass decoding problem~\cite{jin2026adaptingknowledgegraphsbehavior}. Its overhead comes from two sources: semantic preprocessing and MCTS-based trajectory exploration. The preprocessing stage is relatively moderate-scale, as shown in Table~\ref{tab:information-costs}, requiring only 5.1 calls and 2,803.2 tokens per query on LogicCat, and 6.2 calls and 3,424.7 tokens on Archer. Therefore, the dominant cost does not come from schema understanding or semantic extraction, but from repeatedly expanding, evaluating, and selecting candidate reasoning trajectories during search.

Table~\ref{tab:inference-costs} compares IESR with representative Text-to-SQL systems on LogicCat. With $N_{\text{rollout}}=32$, IESR achieves 21.66 EX using 43.8 average calls and 80.7K tokens per query. The token cost is higher than most baselines because each rollout is not a minimal SQL-generation call. Instead, a rollout contains richer intermediate context, including decomposed task information, schema-grounded reasoning states, candidate SQL construction, execution-oriented feedback, and trajectory-level verification signals. This makes the average token cost per call substantially larger than methods that mainly rely on direct generation, retrieval-augmented prompting, or shallow self-consistency. Compared with Chase-SQL, IESR uses fewer calls but a comparable token budget, which suggests that IESR shifts the cost from call frequency to context-rich trajectory construction. Compared with Alpha-SQL, IESR obtains higher EX under a similar search-oriented setting, but consumes more tokens because its reasoning traces and verifier prompts preserve more state information across candidate trajectories.

Table~\ref{tab:inference-rollout} further shows that increasing $N_{\text{rollout}}$ does not translate into monotonic accuracy improvement. Accuracy peaks at $N_{\text{rollout}}=32$ with 21.66 EX, while larger rollout budgets from $N=40$ to $N=64$ consume more calls and tokens but do not yield consistent gains. This pattern indicates a saturation effect in the search space: after a moderate number of rollouts, additional trajectories are more likely to revisit semantically similar query plans, explore low-quality branches, or amplify noisy verifier preferences. The cost increase is also sublinear with respect to the nominal rollout budget. For example, increasing the budget from $N=32$ to $N=64$ doubles the rollout limit, but the average token usage rises only from 80.7K to 95.8K. This is because the rollout number is an upper bound: pruning, early termination, invalid-branch filtering, and shared preprocessing reduce the effective amount of newly generated context. However, the reduced marginal cost does not imply improved efficiency, since the additional tokens bring little or no accuracy gain.

We therefore select $N_{\text{rollout}}=32$ as the default setting. It provides the best execution accuracy in the rollout sweep and represents the best trade-off between search depth and inference budget. In practice, the wall-clock cost can be further reduced through batched rollout decoding, prompt caching, trajectory reuse, and parallel execution across GPUs, while preserving the same logical search procedure.

\begin{table}[htbp]
\centering
\small
\resizebox{\columnwidth}{!}{%
\begin{tabular}{lccc}
\toprule
      Method      &   EX(\%)   & Avg.\ calls & Avg.\ total tokens \\
\midrule
Din-SQL& 10.31 & 3.2 & 2.5K  \\
DAIL-SQL& 10.73 & 3.4 & 2.3K  \\
CHESS& 16.16 & 20.2 & 7.5K  \\
Alpha-SQL& 18.32 & 61.1 & 45.2K  \\
Chase-SQL& 21.92 & 125.2 & 83.1K  \\
Deepeye-SQL& 22.81 & 21.0 & 43.1K  \\
LinkAlign& 19.56 & 42.2 & 19.3K  \\
DTS-SQL& 14.12 & 9.2 & 11.4K  \\
IESR(N=32)& 21.66 & 43.8 & 80.7K  \\
\bottomrule

\end{tabular}
}
\caption{Inference costs of different methods on LogicCat. We report execution accuracy, average model calls, and average total tokens per question.}
\label{tab:inference-costs}
\end{table}

\begin{table}[htbp]
\centering
\small
\resizebox{\columnwidth}{!}{%
\begin{tabular}{lccc}
\toprule
      Method      &   EX(\%)   & Avg.\ calls & Avg.\ total tokens \\
\midrule

IESR, N=4& 20.68 & 11.7 & 22.1K  \\
IESR, N=8& 20.68 & 23.9 & 25.9K  \\
IESR, N=16& 20.40 & 33.2 & 45.7K  \\
IESR, N=32& 21.66 & 43.8 & 80.7K  \\
IESR, N=40& 21.12 & 48.9 & 83.4K \\
IESR, N=48& 20.21 & 54.4 & 89.6K \\
IESR, N=56& 21.39 & 60.7 & 91.9K \\
IESR, N=64& 21.15 & 63.9 & 95.8K \\
\bottomrule

\end{tabular}
}
\caption{Effect of the rollout budget on IESR inference cost and execution accuracy on LogicCat. Larger rollout budgets increase the search budget, but accuracy saturates after $N_{\text{rollout}}=32$.}
\label{tab:inference-rollout}
\end{table}

\begin{table}[htbp]
\centering
\small
\begin{tabular}{lcc}
\toprule
                 & LogicCat & Archer \\
\midrule
Avg.\ calls      & 5.1    & 6.2  \\
Avg.\ total tokens & 2,803.2  & 3,424.7 \\
\bottomrule
\end{tabular}
\caption{Information understanding costs of IESR on LogicCat and Archer. We report the average number of moderate-scale model calls and generated tokens required for semantic extraction and schema-aware preprocessing per query.}
\label{tab:information-costs}
\end{table}

\section{Fine-grained Error Analysis}
\label{sec:errorAnalysis}

\begin{figure}[ht]
        \centering
        \includegraphics[width=\linewidth]{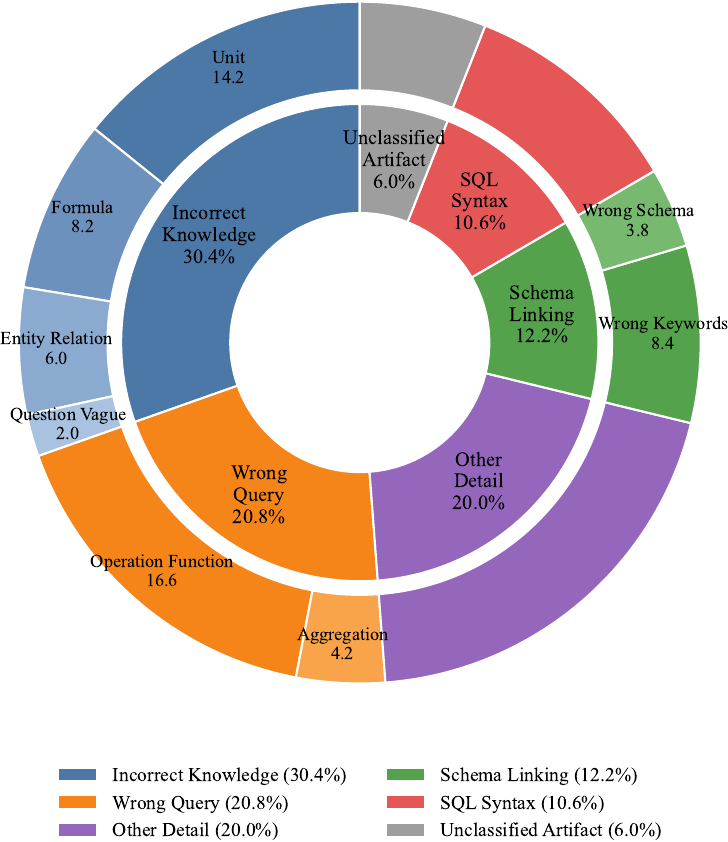}
        \caption{Distribution of Errors on Sampled Set.}
        \label{fig:logicCat_error}
\end{figure}

We conduct a fine-grained failure analysis of the IESR framework on 500 randomly sampled LogicCat examples. Figure~\ref{fig:logicCat_error} reports the error distribution. The analysis reveals that the remaining failures are not dominated by a single source; instead, execution accuracy is constrained by a coupled decision chain involving schema grounding, value normalization, query-plan construction, domain-formula application, and answer-schema preservation.

\begin{itemize}
    \item \textbf{Knowledge-Grounded Reasoning.}
    \textit{Incorrect Knowledge} accounts for the largest fraction of errors (30.4\%). These errors arise when the model fails to apply the domain operation required by the question, including unit conversion, formula selection, physical-quantity interpretation, entity-relation reasoning, and vague intent resolution. Cases~C and D illustrate this issue: the generated SQL is executable, but the computed denotation is wrong because the model uses an incorrect statistical definition or an invalid physical constant. This suggests that SQL executability alone is insufficient for these tasks; the verifier must also judge whether the generated program implements the intended domain computation.
    
    \item \textbf{Compositional Query Planning.}
    \textit{Wrong Query} errors constitute 20.8\% of the sampled failures. In these cases, the model often selects a globally wrong computational strategy while still producing syntactically valid SQL. Typical patterns include over-restricting the query scope, using an inappropriate aggregation, or replacing a required derived computation with a counting objective. Cases~B and F demonstrate that the model can partially ground the relevant table or value but still fail to preserve the target answer semantics. This indicates that the main difficulty is not always locating the right schema element; it is often maintaining the correct transformation sequence from question intent to executable query.

    \item \textbf{Tuple-Schema Preservation.}
    \textit{Other Detail} errors account for 20.0\% of the failures. These errors are local in form but decisive under strict execution matching. Case~H shows a representative projection failure: the join path, arithmetic expression, and row cardinality are largely correct, yet the prediction omits the required key column, causing a tuple-schema mismatch. This type of error suggests that evaluation-sensitive details, such as projected columns, output arity, numeric safeguards, and result ordering assumptions, should be explicitly represented in search and verification.

    \item \textbf{Schema--Dialect Grounding.}
    \textit{Schema Linking} errors (12.2\%) include hallucinated tables, wrong schema elements, and incorrect keyword-to-column grounding, as shown in Case~A. \textit{SQL Syntax} errors (10.6\%) are typically caused by dialect-level constraints, such as unquoted reserved identifiers, as shown in Case~G. Although these categories are smaller than reasoning and planning errors, they remain important because they cause execution failure before denotation can be compared.
\end{itemize}

Overall, the error profile suggests that future improvements should not only strengthen schema retrieval or syntax repair, but also introduce verification signals that evaluate whether a candidate SQL query preserves the intended denotation. In particular, robust text-to-SQL agents for LogicCat-style tasks need to check four properties jointly: whether the schema elements are valid, whether categorical values are normalized to database literals, whether the query plan implements the intended domain computation, and whether the projected answer schema matches the benchmark target.

\begin{figure}[htb]
     \centering
     \includegraphics[width=\linewidth]{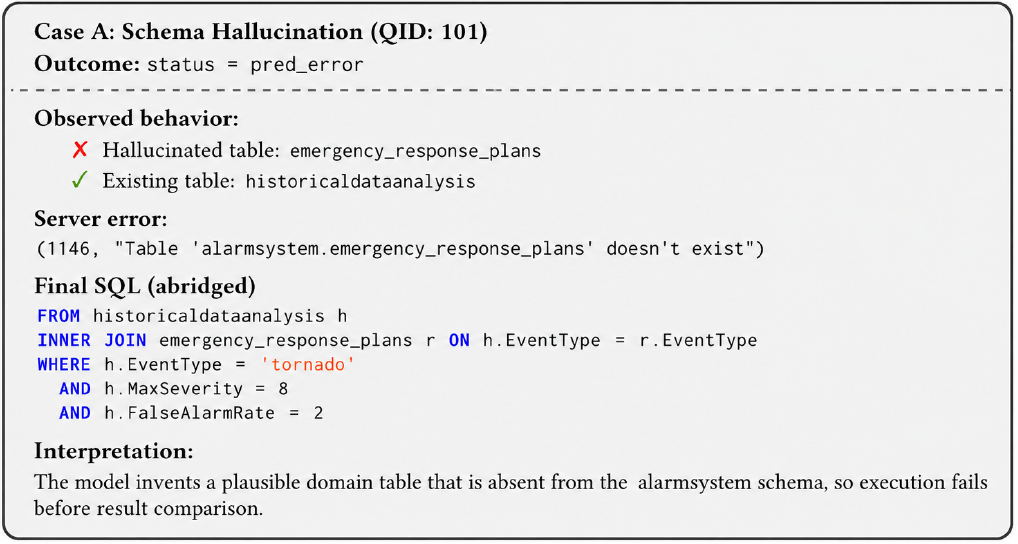}
     \caption{Case A: schema hallucination caused by a non-existent generated table.}
     \label{fig:case-schema-hallucination}
\end{figure}

\begin{figure}[htb]
     \centering
     \includegraphics[width=\linewidth]{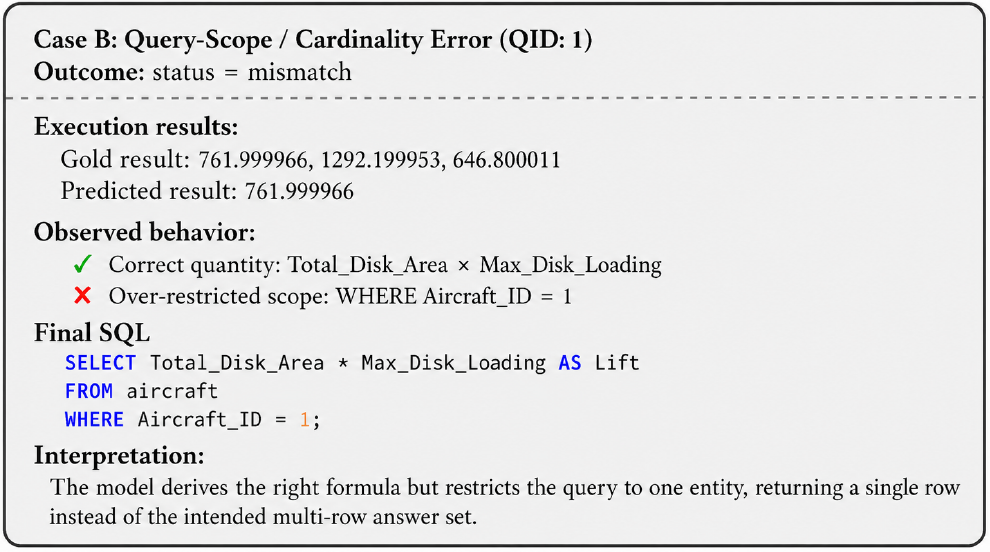}
     \caption{Case B: query-scope error where a multi-row answer is restricted to one entity.}
     \label{fig:case-query-scope}
\end{figure}

\begin{figure}[htb]
     \centering
     \includegraphics[width=\linewidth]{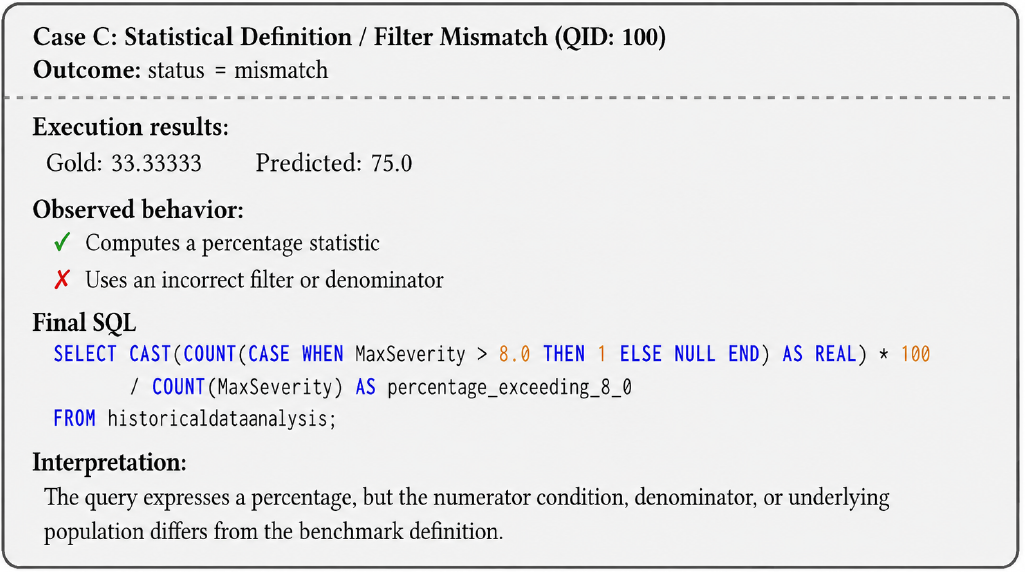}
     \caption{Case C: denotation mismatch caused by an incorrect statistical filter or denominator.}
     \label{fig:case-statistical-filter}
\end{figure}

\begin{figure}[htb]
     \centering
     \includegraphics[width=\linewidth]{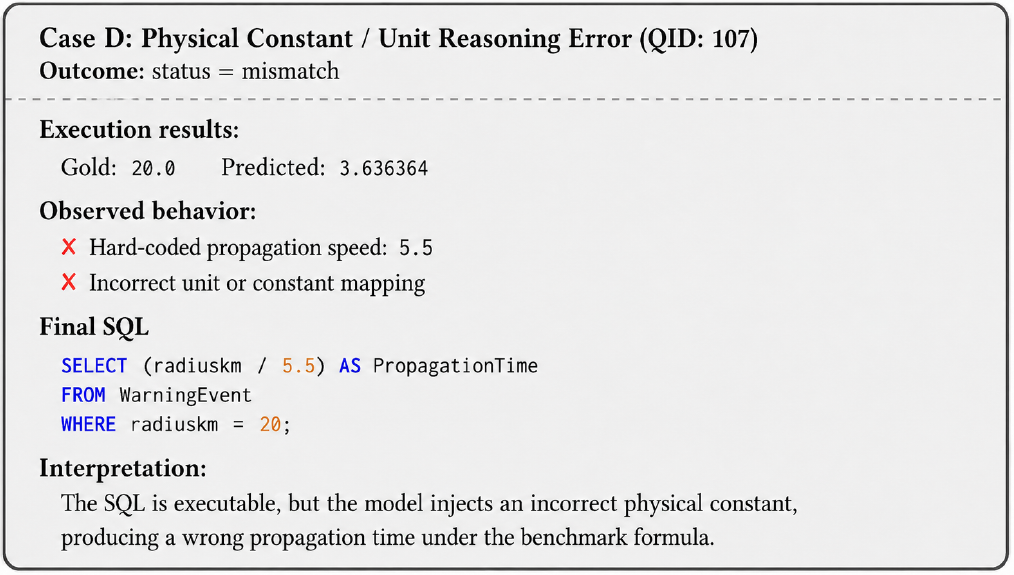}
     \caption{Case D: unit-reasoning error caused by an incorrect physical constant.}
     \label{fig:case-unit-constant}
\end{figure}

\begin{figure}[htb]
     \centering
     \includegraphics[width=\linewidth]{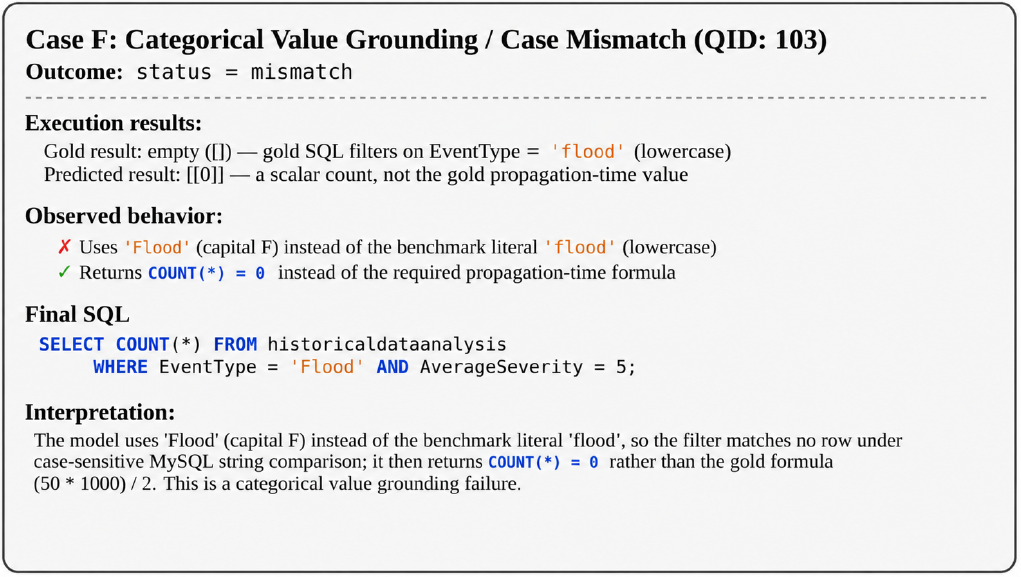}
     \caption{Case F: categorical value grounding error caused by case-sensitive value mismatch.}
     \label{fig:case-value-grounding}
\end{figure}

\begin{figure}[htb]
     \centering
     \includegraphics[width=\linewidth]{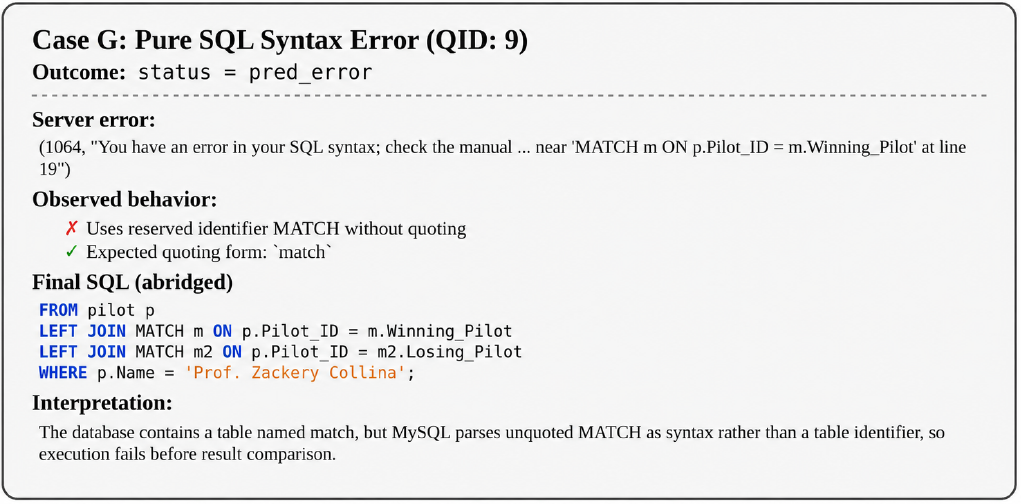}
     \caption{Case G: SQL dialect error caused by an unquoted reserved identifier.}
     \label{fig:case-syntax-dialect}
\end{figure}

\begin{figure}[htb]
     \centering
     \includegraphics[width=\linewidth]{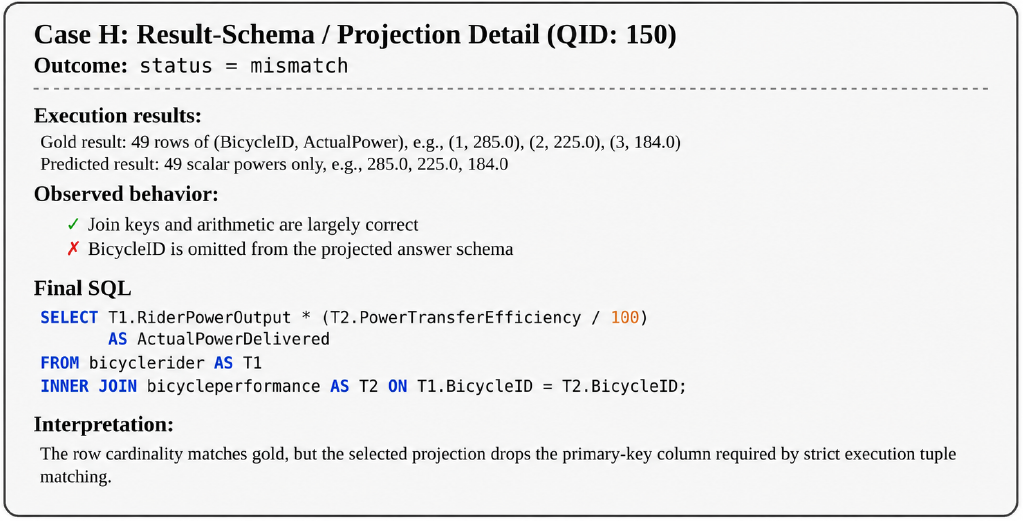}
     \caption{Case H: projection error caused by omitting the required key column.}
     \label{fig:case-projection-detail}
\end{figure}

\section{Algorithm}
\label{sec:nodes}
\subsection{Action Nodes}
Inspired by human reasoning processes, shown in Table~\ref{fig:ActionA15}, we design a diverse action space consisting of five reasoning actions: 
\begin{itemize} 
    \item \textbf{Equation Analysis:} Building on prior work~\cite{mao2025steinersqlgraphguidedmathematicalreasoning} that emphasizes the explicit expression of formulas, the action space is introduced to enable their explicit representation.
    \item \textbf{Schema Selection:} Databases typically contain large and complex schema, while individual queries often involve only a small subset of them. This discrepancy poses significant challenges for SQL generation. Prior work~\cite{li2025alphasql,talaei2024chesscontextualharnessingefficient} has addressed this issue by designing schema selection mechanisms to identify relevant schema components, and we similarly leverage chain-of-thought reasoning to select the most relevant columns.
    \item \textbf{Identify Columns:} Database schema often contain ambiguous fields such as \textit{ids, Keys, Columns} and special function like \textit{POWER(x, 2)} rather than \textit{x * x} and so on, while specific queries refer to concrete numerical values, leading to ambiguous semantic references during SQL generation. Prior work~\cite{li2025deepeyesqlsoftwareengineeringinspiredtexttosqlframework} and existing datasets~\cite{lei2025spider,wretblad2024understandingeffectsnoisetexttosql} have highlighted this issue. The purpose of this dynamic space is to leverage semi-structured information extracted during the information extraction stage to identify such ambiguous fields and explicitly resolve their semantic references. 
    \item \textbf{Entity Extraction:} This step extracts relevant schema information based on entities, relations, and formulas identified from the semi-structured information~\cite{liu2025knowledge}. A text-to-SQL system is required not only to accurately capture filtering conditions and user intent, but also to ensure the correctness of the extracted SQL data. 
    \item \textbf{SQL Generation:} SQL generation is the core component of Text-to-SQL systems~\cite{pourreza2025chasesql}. This method particularly excels at handling nested queries. We incorporate this strategy into our reasoning action space. 
    \item \textbf{SQL Revision:} We observe that many errors from CHASE~\cite{pourreza2025chasesql} and CHESS~\cite{talaei2024chesscontextualharnessingefficient} are due to the difficulty of the model in understanding implicit physical formulas and domain-specific background knowledge. This action requires the model to list all known conditions and explicitly express all relevant formulas, units, key fields, and information, and then correct the SQL query accordingly. 
\end{itemize} 
The search process consists of four steps:

\begin{itemize}
  \item \textbf{Selection}: Starting from the root node, actions are selected downward along the tree based on the UCT values until an unexplored node or a terminal node is encountered.
  \item \textbf{Expansion}: Valid actions are expanded at the selected node to generate new child nodes.
  \item \textbf{Simulation}: From the newly expanded node, actions are sampled randomly and expansion continues until a terminal node is reached.
  \item \textbf{Backpropagation}: The reward of the terminal node is computed and propagated back along the path, updating the statistics of all nodes encountered.
\end{itemize}

After multiple \(N_{\text{rollout}}\) rollouts, all complete reasoning trajectories reach terminal nodes are collected, forming a candidate set of SQL queries for subsequent selection and verification.

At each step, the MCTS selects an action from this space and generates the next reasoning step based on the current state, with the aim of maximizing the reasoning capability of \(LLM_1\). 

\subsection{Module Algorithm}
\label{sec:algorithm}
The proposed framework consists of three stages, each summarized in the appended algorithms. 
First, \textit{Information Understanding with Rule-Guided Verification Algorithm} converts a natural language query into structured elements (intent, entities, relations, numbers, units, operators) and filters them with a rule base to ensure semantic consistency. 
Second, \textit{the Chain-of-Thought Reasoning Algorithm based on MCTS} explores possible SQL construction trajectories, where the search tree is expanded through predefined actions, and candidate paths are simulated and rewarded according to execution results and consistency checks. 
Finally, \textit{Trajectory Selection with Mutual Reasoning Consistency Algorithm} re-evaluates these candidates with a secondary model and scores them based on execution accuracy, discriminator confidence, 
and mutual consistency, selecting the most reliable SQL trajectory. 
This pipeline integrates information understanding, search-based reasoning, and consistency verification, which is particularly suited for datasets that require complex numeric or physical computations, such as \textsc{LogicCat} or \textsc{Archer}.

\begin{algorithm*}[htbp]
\caption{Information Understanding with Rule-Guided Verification}
\label{alg:info_extract}
\begin{algorithmic}[1]
\Require Natural language question $q$, database schema $S=(T,C)$, candidate column set $c_q\subseteq C$, rule base $\mathcal{B}=\{R_j\}$, LLM $M_1$, thresholds $\delta_{\text{match}},\tau$
\Ensure High-quality relation set $\mathcal{R}_{\text{high}}$
\State Inject $(q,S)$ into prompt template
\State Use $M_1$ to generate preliminary results: $R=(i,r,E,R,N,U,P)$
\State $\mathcal{R}_{\text{init}}\gets \text{extract}(R)$
\State $\mathcal{R}_{\text{cand}}\gets\emptyset$
\For{each $r_i\in\mathcal{R}_{\text{init}}$}
  \For{each $R_j\in\mathcal{B}$}
    \If{$\text{sim}(r_i,\mathrm{Uni}_j,\mathrm{Equt}_j)>\delta_{\text{match}}$}
      \State $\mathcal{R}_{\text{cand}}\gets \mathcal{R}_{\text{cand}}\cup\{r_i\}$
      \State \textbf{break}
    \EndIf
  \EndFor
\EndFor
\State $\mathcal{R}_{\text{high}}\gets\emptyset$
\For{each $r_i\in\mathcal{R}_{\text{cand}}$}
  \State $P_i\gets \text{Plan}(r_i,\mathcal{B})$
  \State $s_i\gets \text{Executor}(r_i,P_i)$
  \If{$s_i>\tau$}
    \State $\mathcal{R}_{\text{high}}\gets \mathcal{R}_{\text{high}}\cup\{r_i\}$
  \EndIf
\EndFor
\State \Return $\mathcal{R}_{\text{high}}$
\end{algorithmic}
\end{algorithm*}

\begin{algorithm*}[htbp]
\caption{MCTS-based CoT Reasoning for SQL Generation}
\label{alg:mcts_cot}
\begin{algorithmic}[1]
\Require Query $x$, target LLM $M$, action set $\mathcal{A}=\{\mathrm{A1},\dots,\mathrm{A6}\}$, rollout budget $N_{\text{rollout}}$, exploration constant $c$, sampling temperature $T_{\text{samp}}$, consistency vote size $K$, database $D$
\Ensure Candidate trajectory set $\mathcal{T}_{\text{cand}}$ with terminal rewards
\State Initialize search tree $\mathcal{T}$ with root node $v_0$ representing $x$
\State Initialize visit counts $N(v,a)$ and cumulative values $Q(v,a)$ for all $(v,a)$; $\mathcal{T}_{\text{cand}}\gets \emptyset$
\For{$m=1$ to $N_{\text{rollout}}$} 
  \State $v \gets v_0$;\quad $\textit{path}\gets [\,]$ 
  \Comment{Selection}
  \While{\textbf{not} terminal$(v)$ \textbf{and} $v$ is fully expanded}
      \State $a^\star \gets \arg\max_{a\in\mathcal{A}}
      \left(\frac{Q(v,a)}{\max(1,N(v,a))} +
      c\sqrt{\frac{\ln \max(1,N(v))}{\max(1,N(v,a))}}\right)$
      \State $\textit{path}.\mathrm{append}\big((v,a^\star)\big)$;\quad $v \gets \textsc{Step}(v,a^\star, M)$
  \EndWhile
  \Comment{Expansion}
  \If{\textbf{not} terminal$(v)$}
    \State Select an untried valid action $a$ at $v$
    \State $u \gets \textsc{Step}(v,a,M)$; add edge $(v,a)\to u$; initialize $N(u,\cdot),Q(u,\cdot)$
    \State $\textit{path}.\mathrm{append}\big((v,a)\big)$;\quad $v \gets u$
  \EndIf
  \Comment{Simulation (rollout)}
  \State $(\hat{t}, \textit{terminal}) \gets \textsc{Simulate}(v, M, \mathcal{A}, T_{\text{samp}})$
  \State $r_t \gets 0$
  \If{\textit{terminal} = \textbf{true}}
    \State $r_t \gets \textsc{EvaluateReward}(\hat{t}, x, D, K)$
    \State $\mathcal{T}_{\text{cand}} \gets \mathcal{T}_{\text{cand}} \cup \{(\hat{t}, r_t)\}$
  \EndIf
  \Comment{Backpropagation}
  \State \textsc{Backpropagate}($\textit{path}$, $r_t$)
\EndFor
\State \Return $\mathcal{T}_{\text{cand}}$
\vspace{0.25em}

\Function{Step}{$v, a, M$} 
  \State Apply action $a$ to extend the current reasoning state at $v$ using model $M$
  \State \Return new node $u$ with updated partial trajectory
\EndFunction

\Function{Simulate}{$v, M, \mathcal{A}, T_{\text{samp}}$}
  \While{\textbf{not} terminal$(v)$ \textbf{and} depth limit not reached}
    \State Sample $a$ from a rollout policy (e.g., uniform or softmax over UCT scores with temperature $T_{\text{samp}}$)
    \State $v \gets \textsc{Step}(v,a,M)$
  \EndWhile
  \State \Return (final trajectory $\hat{t}$ from root$\to v$, terminal$(v)$)
\EndFunction

\Function{EvaluateReward}{$\hat{t}, x, D, K$} 
  \State Extract terminal SQL $\hat{y}$ from $\hat{t}$
  \If{$\textsc{Execute}(\hat{y},D)$ fails}
    \State \Return $0$
  \EndIf
  \State Sample $K$ additional terminal SQL queries $\{\hat{y}_j\}_{j=1}^{K}$ under the same terminal context
  \State Execute all; let $w \gets \frac{1}{K}\sum_{j=1}^{K}\mathbf{1}\big[\textsc{Execute}(\hat{y},D)=\textsc{Execute}(\hat{y}_j,D)\big]$
  \State \Return $r_t = w$
\EndFunction

\Function{Backpropagate}{$\textit{path}, r_t$}
  \For{each $(v,a)$ in $\textit{path}$}
    \State $N(v,a) \gets N(v,a) + 1$;\quad $Q(v,a) \gets Q(v,a) + r_t$
  \EndFor
\EndFunction
\end{algorithmic}
\end{algorithm*}

\begin{algorithm*}[htbp]
\caption{Trajectory Selection with Mutual Reasoning Consistency}
\label{alg:traj_select}
\begin{algorithmic}[1]
\Require Candidate trajectory set $\mathcal{T}$ from $LLM_{1}$, discriminator $LLM_{2}$, weights $\alpha, \beta, \gamma$
\Ensure Final selected SQL trajectory $t^{*}$

\State $\mathcal{T}_{\text{cand}} \gets \emptyset$
\For{each trajectory $t = x \oplus s_1 \oplus \cdots \oplus s_d \in \mathcal{T}$}
  \State Randomly choose $i<d$, mask $s_i,\dots,s_d$
  \State Feed $(x \oplus s_1 \oplus \cdots \oplus s_{i-1})$ into $LLM_{2}$ to generate completion $\hat{t}$
  \If{$\hat{t}$ semantically consistent with $t$}
    \State $\mathcal{T}_{\text{cand}} \gets \mathcal{T}_{\text{cand}} \cup \{t\}$
  \EndIf
\EndFor

\For{each $t \in \mathcal{T}_{\text{cand}}$}
  \State Compute $Exec(t)$ 
  \State Compute $DiscConf(t)$ 
  \State Compute $ConsVote(t)$ 
  \State $Score(t) \gets \alpha \cdot Exec(t) + \beta \cdot DiscConf(t) + \gamma \cdot ConsVote(t)$
\EndFor

\State $t^{*} \gets \arg\max_{t \in \mathcal{T}_{\text{cand}}} Score(t)$
\State \Return $t^{*}$
\end{algorithmic}
\end{algorithm*}

\section{Experiments Hyperparameter}
\label{sec:impl_config}
Table~\ref{tab:implementation_config} reports the MCTS and local decoding parameters used in our experiments.

\begin{table}[t]
\centering
\small
\setlength{\tabcolsep}{4pt}
\renewcommand{\arraystretch}{1.08}
\resizebox{\columnwidth}{!}{
\begin{tabular}{l l l}
\toprule
Category & Parameter & Value \\
\midrule
MCTS & reasoning\_variant & full\_iesr \\
MCTS & enable\_sql\_revision & true \\
MCTS & max\_rollout\_steps & 16 \\
MCTS & max\_depth & 16 \\
MCTS & exploration\_constant & 1.414 \\
MCTS & random\_seed & 42 \\
MCTS & n\_processes & 10 \\
\midrule
LLM decoding & n & 3 \\
LLM decoding & top\_p & 0.8 \\
LLM decoding & max\_tokens & 1024 \\
LLM decoding & temperature & 0.8 \\
LLM decoding & n\_strategy & single \\
\bottomrule
\end{tabular}
}
\caption{MCTS and local LLM decoding configuration used in our experiments.}
\label{tab:implementation_config}
\end{table}

\section{Hyperparameter Sensitivity Analysis}

\label{sec:weight_analysis}

We analyze the sensitivity of two manually specified components in IESR:
the trajectory selection weights $(\alpha,\beta,\gamma)$ and the schema
filtering threshold $\tau$. These experiments are designed to examine
whether the final performance depends on a narrow heuristic configuration.

\begin{table}[t]
\centering
\scriptsize
\setlength{\tabcolsep}{2.5pt}
\renewcommand{\arraystretch}{1.05}
\resizebox{\columnwidth}{!}{
\begin{tabular}{l c c}
\toprule
Setting & EX (\%) & Valid SQL (\%) \\
\midrule
$\gamma > 0$ configs
& $18.7\sim21.5$
& $54.2\sim57.0$ \\
$\gamma = 0$ configs
& $18.8\sim21.0$
& $54.0\sim57.2$ \\
Full 21-tuple grid
& $18.7\sim21.5$
& $54.1\sim57.2$ \\
\bottomrule
\end{tabular}
}
\caption{Sensitivity summary of trajectory-selection weights on 500 LogicCat development queries using Qwen2.5-Coder-7B-Instruct. Candidate trajectories are fixed across all settings.}
\label{tab:weight_sensitivity_summary}
\end{table}

\paragraph{Trajectory selection weights.}
We first study the weighting coefficients in the trajectory scoring function:
\begin{equation*}
\small
\begin{aligned}
\text{Score}(t)
&= \alpha \cdot \text{Exec}(t)
 + \beta \cdot \text{DiscConf}(t)
 + \gamma \cdot \text{ConsVote}(t).
\end{aligned}
\end{equation*}

Here, $\text{Exec}(t)$ denotes the execution based signal,
$\text{DiscConf}(t)$ denotes the discriminator confidence or syntactic
validity proxy, and $\text{ConsVote}(t)$ denotes the trajectory consistency
signal derived from peer agreement. The execution signal provides the main
selection criterion, while the two consistency signals serve as auxiliary
evidence for suppressing unstable reasoning trajectories.

\begin{table}[htp]
\centering
\small
\renewcommand{\arraystretch}{1.05}
\begin{tabular}{c c c c c c}
\toprule
$\alpha$ & $\beta$ & $\gamma$ & Samples & EX (\%) & Valid SQL (\%) \\
\midrule
0.0 & 0.0 & 1.0 & 500 & 19.3 & $\sim 54.3$ \\
0.0 & 0.2 & 0.8 & 500 & 19.4 & $\sim 55.8$ \\
0.0 & 0.4 & 0.6 & 500 & 19.6 & $\sim 55.9$ \\
0.0 & 0.6 & 0.4 & 500 & 20.4 & $\sim 55.7$ \\
0.0 & 0.8 & 0.2 & 500 & 19.0 & $\sim 56.8$ \\
0.0 & 1.0 & 0.0 & 500 & 19.3 & $\sim 54.9$ \\
0.2 & 0.0 & 0.8 & 500 & 21.1 & $\sim 54.2$ \\
0.2 & 0.2 & 0.6 & 500 & 18.7 & $\sim 55.5$ \\
0.2 & 0.4 & 0.4 & 500 & 21.2 & $\sim 56.8$ \\
0.2 & 0.6 & 0.2 & 500 & 20.1 & $\sim 56.8$ \\
0.2 & 0.8 & 0.0 & 500 & 20.2 & $\sim 56.9$ \\
0.4 & 0.0 & 0.6 & 500 & 19.8 & $\sim 54.9$ \\
0.4 & 0.2 & 0.4 & 500 & \textbf{21.5} & $\sim 55.2$ \\
0.4 & 0.4 & 0.2 & 500 & 21.4 & $\sim 57.0$ \\
0.4 & 0.6 & 0.0 & 500 & 21.0 & $\sim 56.8$ \\
0.6 & 0.0 & 0.4 & 500 & 20.8 & $\sim 54.3$ \\
0.6 & 0.2 & 0.2 & 500 & 21.3 & $\sim 54.8$ \\
0.6 & 0.4 & 0.0 & 500 & 19.1 & $\sim 57.2$ \\
0.8 & 0.0 & 0.2 & 500 & 18.7 & $\sim 55.3$ \\
0.8 & 0.2 & 0.0 & 500 & 19.0 & $\sim 56.1$ \\
1.0 & 0.0 & 0.0 & 500 & 18.8 & $\sim 54.0$ \\
\bottomrule
\end{tabular}
\caption{Full grid over trajectory selection weights with step size $0.2$.
All configurations satisfy $\alpha+\beta+\gamma=1$.}
\label{tab:weight_sensitivity_full}
\end{table}

\begin{table}[t]
\centering
\scriptsize
\setlength{\tabcolsep}{2.5pt}
\renewcommand{\arraystretch}{1.05}
\resizebox{\columnwidth}{!}{
\begin{tabular}{c c c c c}
\toprule
$\tau$ & Macro F1 & Macro P & Macro R & Col. Recall \\
\midrule
0.3 & 25.4 & 24.5 & 66.8 & 72.8 \\
0.4 & 30.3 & 39.2 & 69.3 & 71.3 \\
0.5 & 37.9 & 38.6 & 66.3 & 68.3 \\
0.6 & \textbf{51.5} & \textbf{51.3} & \textbf{74.8} & \textbf{76.3} \\
0.7 & 49.2 & 48.9 & 69.0 & 75.0 \\
\bottomrule
\end{tabular}
}
\caption{Sensitivity analysis of the schema filtering threshold $\tau$ on 500 LogicCat development queries. Metrics are computed on retrieved triplets against gold SQL literals.}
\label{tab:tau_sensitivity}
\end{table}

To isolate the effect of these coefficients, we perform offline reranking
over saved candidate trajectories produced by the full IESR pipeline. This
setting keeps the candidate SQL set fixed and changes only the final scoring
weights, introducing no additional LLM calls. Specifically, we use 500
LogicCat development queries with saved rollouts from
Qwen2.5-Coder-7B-Instruct and exhaustively sweep all tuples satisfying
$\alpha,\beta,\gamma \geq 0$ and $\alpha+\beta+\gamma=1$ with step size
$0.2$, resulting in 21 configurations.

Table~\ref{tab:weight_sensitivity_summary} summarizes the sweep results.
The valid SQL rate stays within a narrow range across different weights,
suggesting that these coefficients mainly affect final candidate selection
among generated trajectories rather than syntactic validity. The default
configuration $(\alpha,\beta,\gamma)=(0.4,0.2,0.4)$ achieves the best EX in
this grid. Configurations with $\gamma>0$ are generally competitive,
indicating that trajectory consistency provides useful auxiliary evidence.
Meanwhile, using only one signal, such as $(1.0,0.0,0.0)$ or
$(0.0,0.0,1.0)$, gives lower EX, showing that execution and consistency
signals are complementary.

\paragraph{Schema filtering threshold.}
We further analyze the LSH MinHash threshold $\tau$ used in schema filtering.
This threshold controls the precision recall tradeoff of retrieved
$(\text{table},\text{column},\text{value})$ triplets before downstream
reasoning. We sweep $\tau$ on the same 500 LogicCat development queries using
the Stage 1 pipeline with MySQL execution, local LLM keyword extraction, and
local embedding filters. We report macro averaged precision, recall, and F1
against gold SQL literals, together with macro column recall over gold
columns. Since 56 examples contain no parser extracted literal in the gold
SQL, literal level F1 can be noisy; column recall is therefore included as a
more stable schema grounding indicator.

The best macro F1 is obtained at $\tau=0.6$, which is used as the default
schema filtering threshold in all main experiments. Overall, the sensitivity
analysis shows that IESR benefits from calibrated execution and consistency
signals, while its performance does not rely on a single isolated weight
choice. The schema filtering threshold mainly controls the expected
precision recall tradeoff before MCTS reasoning, and the selected default
corresponds to the best development F1 among the tested values.

\section{Dataset Details}
\label{sec:dataset}
\begin{itemize}
     \item \textbf{LogicCAT~\cite{liu2025logiccatchainofthoughttexttosqlbenchmark}:} A challenging Text-to-SQL benchmark with 4,038 English questions covering multi-domain reasoning including physical knowledge, mathematical logic, common sense reasoning, and hypothetical scenarios.
     \item \textbf{Archer~\cite{zheng-etal-2024-archer}:} It contains 1,042 questions from 20 databases that focus on complex, logical, hypothetical reasoning.
     \item \textbf{BIRD~\cite{wretblad2024understandingeffectsnoisetexttosql}:} BIRD contains more than 12,751 unique questions and 95 databases. We used the development set, which includes 1,534 questions.
     \item \textbf{Spider~\cite{yu2019spiderlargescalehumanlabeleddataset}:} Spider contains 10,181 unique questions in 166 databases. We use its development set of 1,034 questions.
\end{itemize}

\section{Details of Public Method}
\label{sec:pubmethod}
\begin{itemize}
     \item \textbf{DIN-SQL:} DIN-SQL is a Text-to-SQL framework that splits query generation into steps (schema linking, decomposition, intermediate representation, correction) and reaches SOTA results without fine-tuning.
     \item \textbf{DAIL-SQL:} DAIL-SQL is a prompt‐engineering method optimizing how examples and question representations are used in a few shots based on LLM Text-to-SQL.
     \item \textbf{CHESS:} CHESS is a multi-agent LLM framework for text-to-SQL, combining an information retriever, schema selector, candidate generator, and unit tester to prune large schemas, generate and refine SQL, and validate outputs. Reduce token usage and model calls while improving accuracy, achieving strong results on large-scale benchmarks such as BIRD.
     \item \textbf{LinkAlign:} It is a scalable schema linking framework for multi-database Text-to-SQL, combining semantic retrieval and schema grounding to accurately select relevant databases and columns in large, complex settings.
     \item \textbf{DTS-SQL:} DTS-SQL is a two-stage fine-tuning method that splits the Text-to-SQL task into schema linking and SQL generation subtasks, allowing smaller open source LLMs to approach the performance of larger models.
     \item \textbf{Alpha-SQL:} Alpha-SQL is a zero-shot Text-to-SQL framework that uses Monte Carlo Tree Search (MCTS) plus dynamic action proposals from an LLM to iteratively build SQL, with a self-supervised reward function to guide the search, achieving strong results.
     \item \textbf{SQL-O1:} SQL-o1 introduces a self-reward heuristic dynamic search method for Text-to-SQL. It leverages execution-aware feedback to guide iterative refinement of candidate SQL queries, enabling effective search without external annotators or learned critics.
     \item \textbf{CHASE-SQL:} CHASE-SQL~\cite{pourreza2025chasesql} is a multi-path reasoning framework that generates diverse SQL candidates through iterative reasoning and applies preference-based candidate selection to identify high-quality solutions.
    \item \textbf{Deepeye-SQL:} Deepeye-SQL~\cite{li2025deepeyesqlsoftwareengineeringinspiredtexttosqlframework} is an agentic Text-to-SQL framework that leverages iterative exploration and execution-aware feedback to improve SQL generation, particularly for complex database reasoning tasks.

\end{itemize}

\section{Prompt}
\label{sec:prompt}
\begin{figure*}[h]
  \centering
  \includegraphics[width=1.0\textwidth]{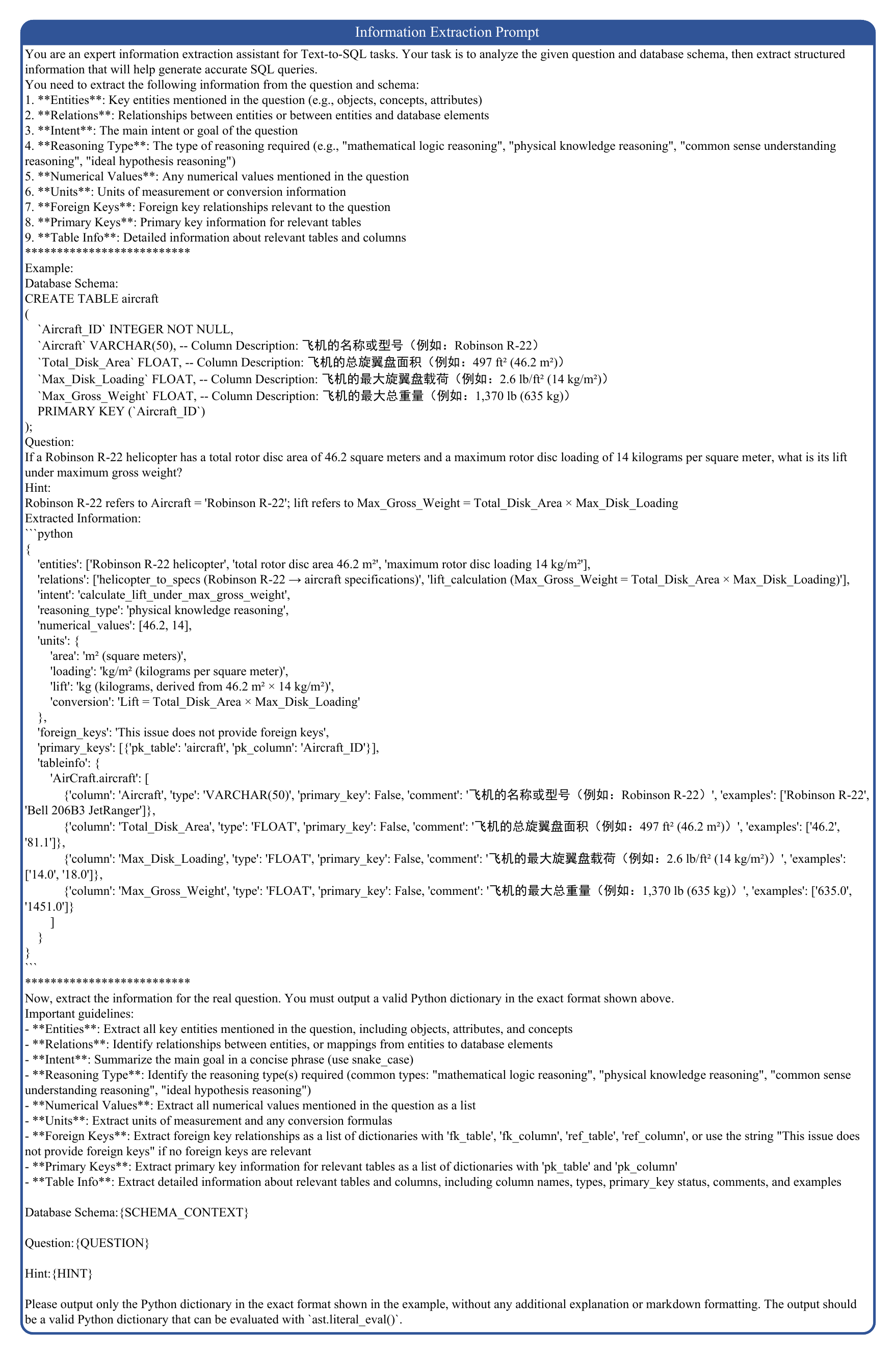}
  \caption{Understanding Prompt including Intent Recognition, Unit Understanding, Relation and Entity Extracting, and Pseudo-Schema Understanding}
  \label{fig:InformationExtractionPrompt}
\end{figure*}

\begin{figure*}[h]
  \centering
  \includegraphics[width=1.0\textwidth]{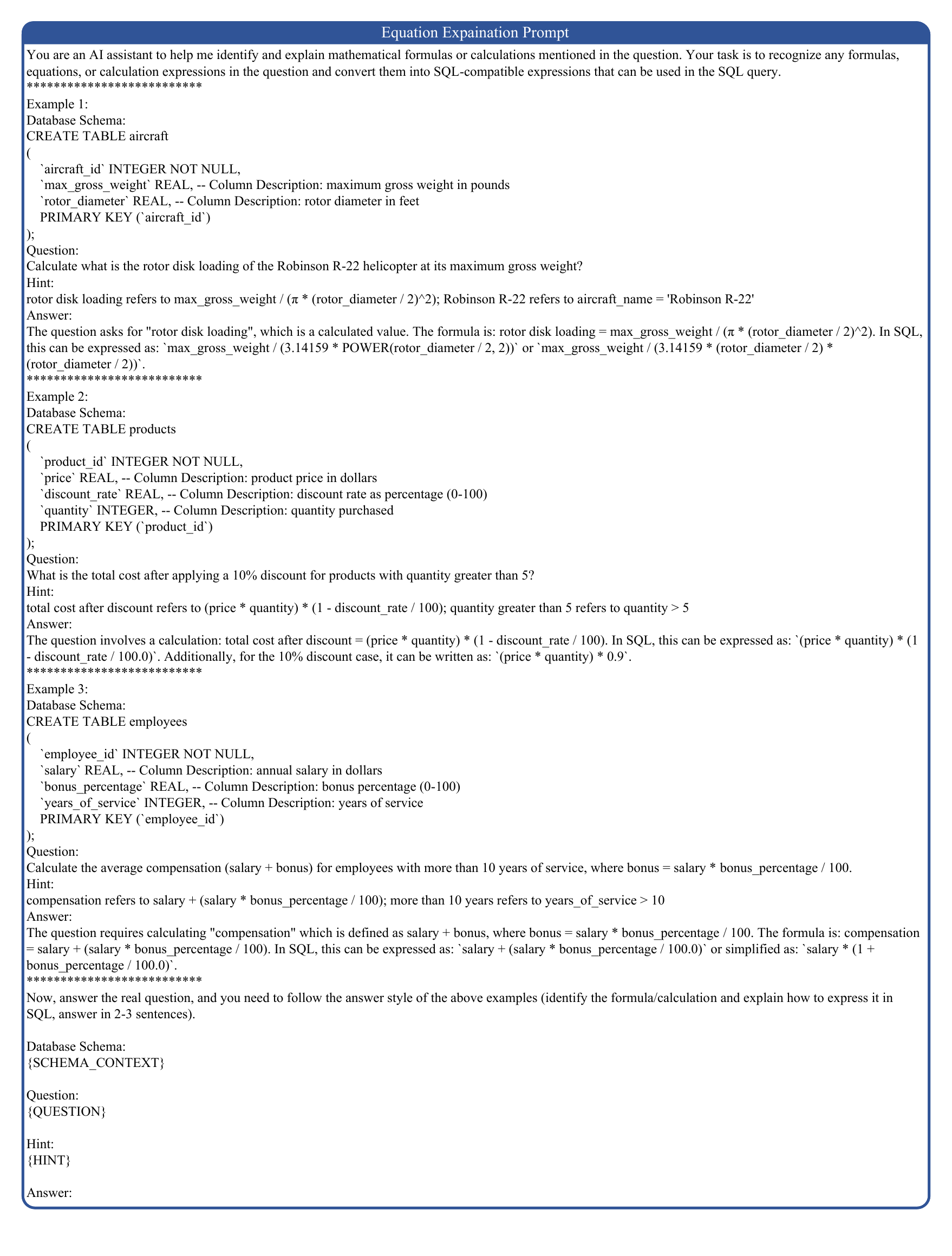}
  \caption{An Example of Equation Explain Prompt.}
  \label{fig:EquationExpainPrompt}
\end{figure*}

\begin{figure*}[h]
  \centering
  \includegraphics[width=1.0\textwidth]{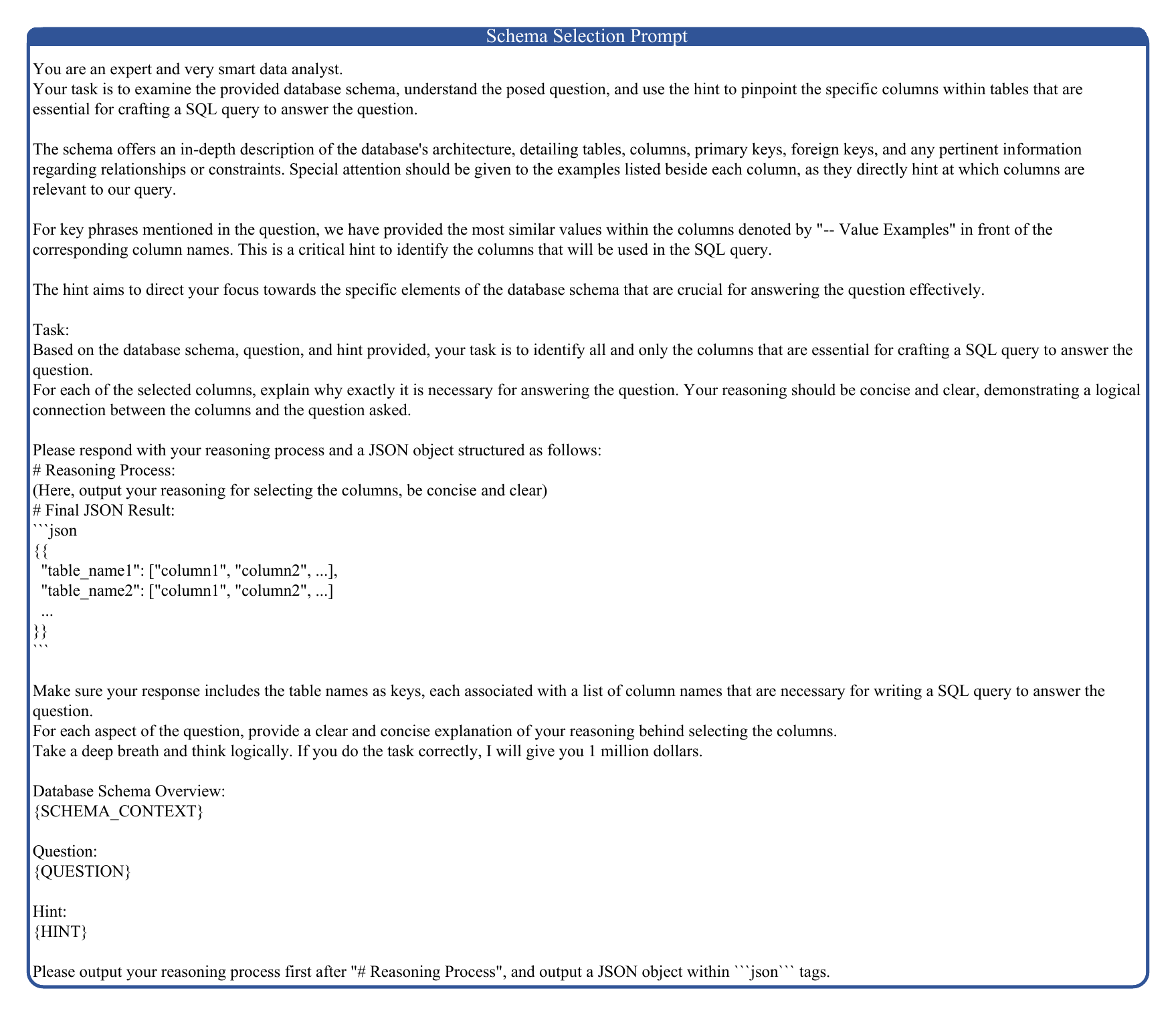}
  \caption{An Example of Schema Selection Prompt.}
  \label{fig:SchemaSelectionPrompt}
\end{figure*}

\begin{figure*}[h]
  \centering
  \includegraphics[width=1.0\textwidth]{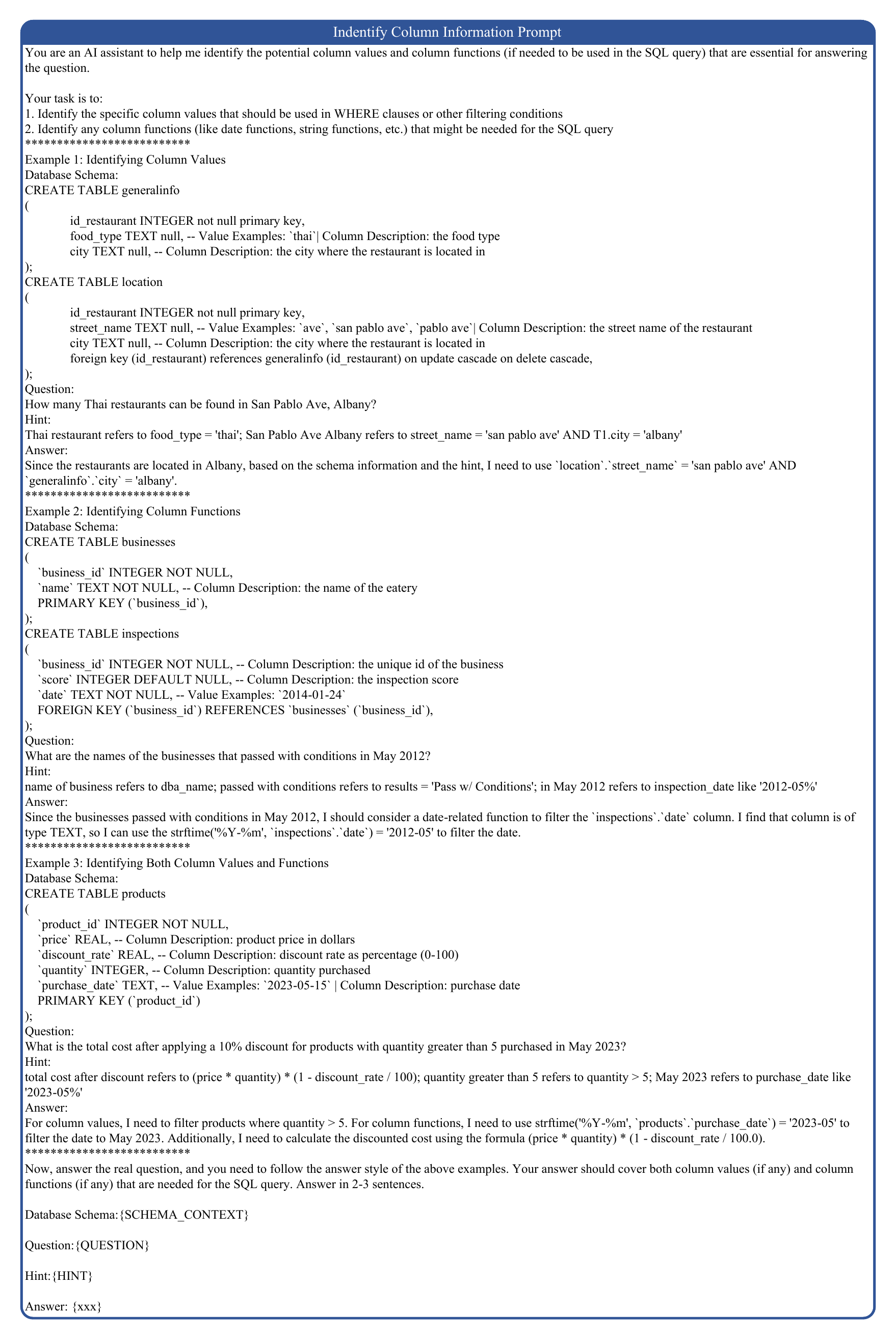}
  \caption{An Example of Identify Column Information Prompt.}
  \label{fig:IndentifyColumnInfomationPrompt}
\end{figure*}

\begin{figure*}[h]
  \centering
  \includegraphics[width=1.0\textwidth]{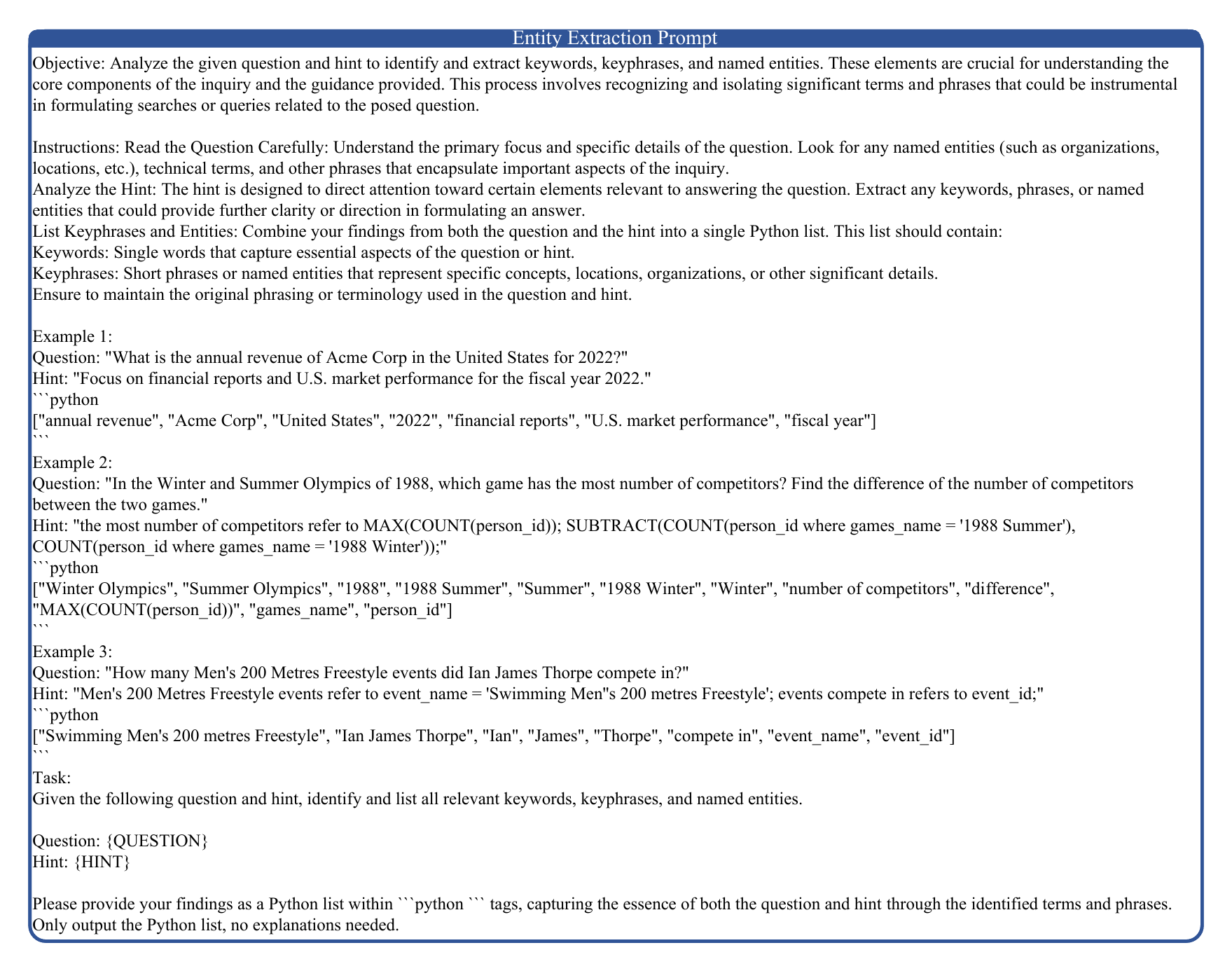}
  \caption{An Example of Entity Extraction Prompt.}
  \label{fig:EntityExtractionPrompt}
\end{figure*}

\begin{figure*}[h]
  \centering
  \includegraphics[width=1.0\textwidth]{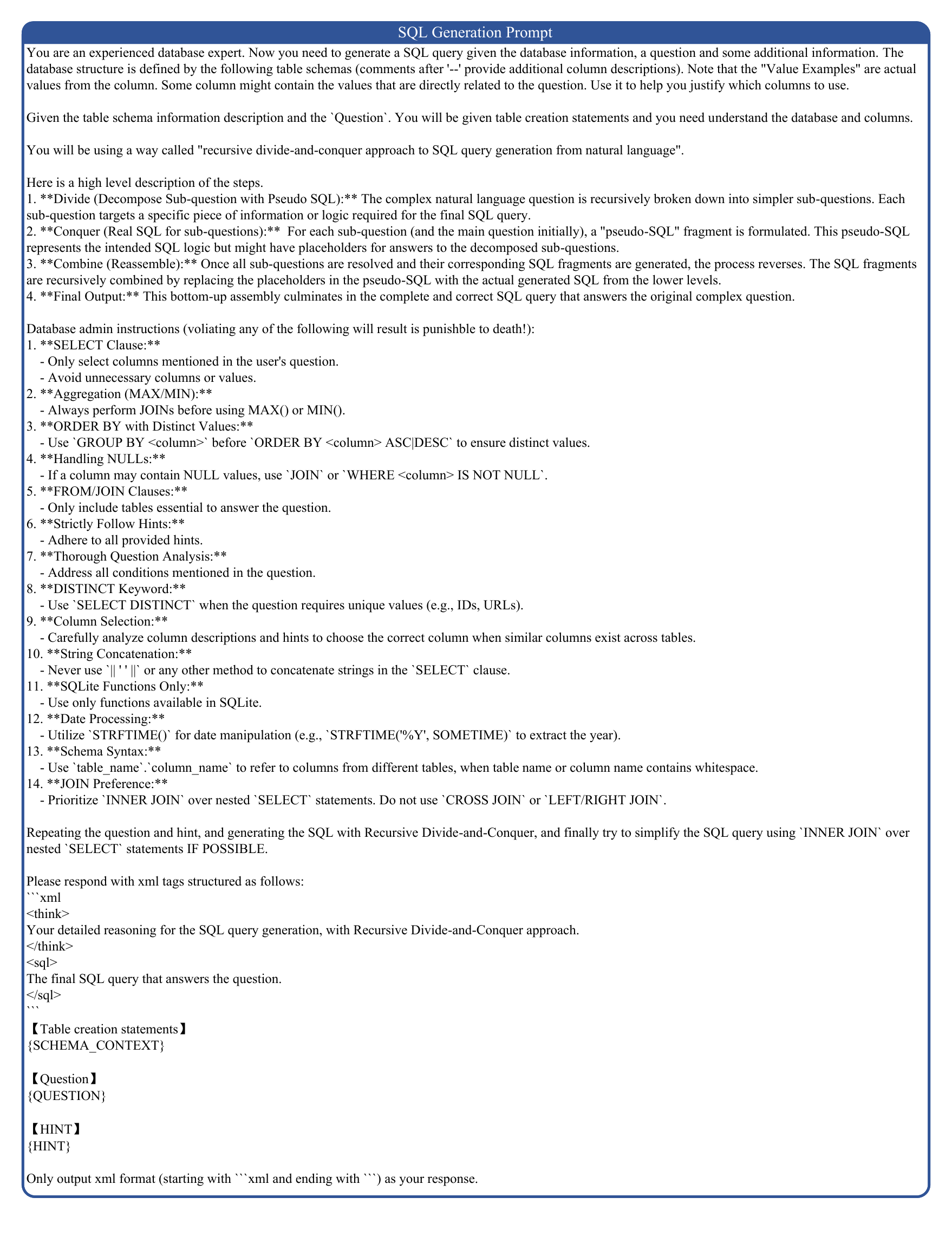}
  \caption{Understanding Prompt including Intent Recognition, Unit Understanding, Relation and Entity Extracting, and Pseudo-Schema Understanding.}
  \label{fig:SqlGenerationPrompt}
\end{figure*}

\begin{figure*}[h]
  \centering
  \includegraphics[width=1.0\textwidth]{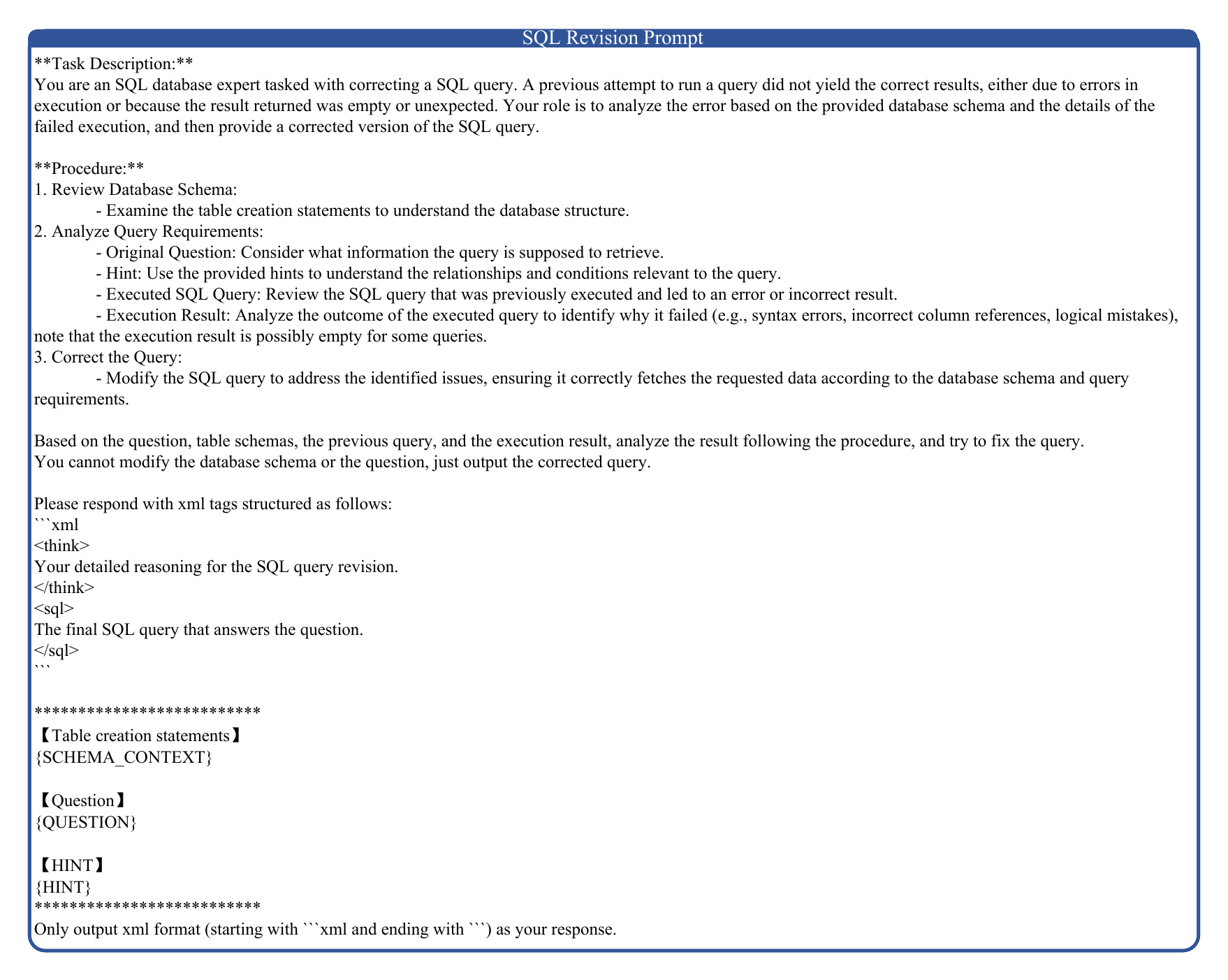}
  \caption{An Example of Revising SQL Reasoning Prompt.}
  \label{fig:SqlRevisionPrompt}
\end{figure*}

In this section, we summarize the prompt templates used in our modular agent, organized as one front-end understanding prompt and six action prompts that constitute the heterogeneous reasoning space. The Information Understanding prompt constructs a structured semantic state from the user question by extracting intent, entities and relations, numerical values and units, and the schema elements that are likely required (Figure \ref{fig:InformationExtractionPrompt}). Based on this intermediate representation, the agent performs reasoning through six specialized actions. The Equation Explain prompt makes implicit formulas and numerical transformations explicit, supporting unit conversion and arithmetic constraints before SQL construction (Figure \ref{fig:EquationExpainPrompt}). The Schema Selection prompt narrows the search space by selecting task-relevant tables and columns from the database schema (Figure \ref{fig:SchemaSelectionPrompt}). To further resolve ambiguous fields, the Identify Column Information prompt disambiguates candidate columns using semantic cues and column-level descriptions/examples (Figure \ref{fig:IndentifyColumnInfomationPrompt}). The Entity Extraction prompt grounds query entities and their associated relations to concrete values or schema mentions, providing reliable constraints for WHERE/JOIN conditions (Figure \ref{fig:EntityExtractionPrompt}). Conditioned on the above structured signals, the SQL Generation prompt synthesizes an executable SQL query, including necessary joins, aggregations, and nested structures (Figure \ref{fig:SqlGenerationPrompt}). Finally, the SQL Revision prompt corrects the draft SQL by diagnosing logical, join, and schema-mismatch errors (optionally using execution feedback when available), and outputs a repaired query (Figure \ref{fig:SqlRevisionPrompt})



\end{document}